\documentclass[10pt,journal,compsoc]{IEEEtran}

\ifCLASSINFOpdf
\else
\usepackage[dvips]{graphicx}
\fi
\usepackage{url}

\usepackage[colorlinks,linkcolor=blue,anchorcolor=blue,
citecolor=blue]{hyperref}
\usepackage{graphicx}
\usepackage{amsmath,amssymb}
\usepackage{booktabs,tabularx,multirow,threeparttable}
\usepackage{float}
\usepackage{subfigure}
\usepackage{amssymb}
\usepackage{subfigure}
\usepackage{amsthm,amsmath,amsfonts,amssymb,bm}

\usepackage{lineno}
\usepackage{hyperref}       
\usepackage{graphicx}

\usepackage{xcolor}         

\definecolor{ForestGreen}{rgb}{0, 0.69, 0.31}
\definecolor{NavyBlue}{rgb}{0, 0.44, 0.75}
\newcommand{\hgreen}[1]{\textcolor{ForestGreen}{\textbf{#1}}} 
\newcommand{\hblue}[1]{\textcolor{NavyBlue}{\textbf{#1}}} 
\definecolor{hblue}{rgb}{0, 0.44, 0.75}
\definecolor{firstBest}{rgb}{0.9, 1, 0.9}
\definecolor{secondBest}{rgb}{1, 0.95, 0.95}

\usepackage{bbding}
\usepackage{makecell}

\ifCLASSOPTIONcompsoc
  \usepackage[nocompress]{cite}
\else
  \usepackage{cite}
\fi

\usepackage[numbers]{natbib} 

\begin{document}
\title{UAVBench and UAVIT-1M: Benchmarking and Enhancing MLLMs for Low-Altitude UAV Vision-Language Understanding}
\author{Yang~Zhan~and~Yuan~Yuan*,~\IEEEmembership{Senior Member,~IEEE}
   
\thanks{Yang Zhan and Yuan Yuan are with the School of Artificial Intelligence, Optics, and Electronics (iOPEN), Northwestern Polytechnical University, Xi'an 710072, China. (e-mail:{zhanyangnwpu@gmail.com})
}
}

\markboth{}%
{Shell \MakeLowercase{\textit{et al.}}: Bare Demo of IEEEtran.cls for IEEE Journals}

\IEEEtitleabstractindextext{
\begin{abstract}
Multimodal Large Language Models (MLLMs) have made significant strides in natural images and satellite remote sensing images. However, understanding low-altitude drone scenarios remains a challenge. Existing datasets primarily focus on a few specific low-altitude visual tasks, which cannot fully assess the ability of MLLMs in real-world low-altitude UAV applications.
Therefore, we introduce \textbf{UAVBench}, a comprehensive benchmark, and \textbf{UAVIT-1M}, a large-scale instruction tuning dataset, designed to evaluate and improve MLLMs’ abilities in low-altitude vision-language tasks. UAVBench comprises 43 test units and 966k high-quality data samples across 10 tasks at the image-level and region-level. UAVIT-1M consists of approximately 1.24 million diverse instructions, covering 789k multi-scene images and about 2,000 types of spatial resolutions with 11 distinct tasks. 
UAVBench and UAVIT-1M feature pure real-world visual images and rich weather conditions, and involve manual verification to ensure high quality. Our in-depth analysis of 11 state-of-the-art MLLMs using UAVBench reveals that open-source MLLMs cannot generate accurate conversations about low-altitude visual content, lagging behind closed-source MLLMs. Extensive experiments demonstrate that fine-tuning open-source MLLMs on UAVIT-1M significantly addresses this gap. Our contributions pave the way for bridging the gap between current MLLMs and low-altitude UAV real-world application demands. (Project page: https://UAVBench.github.io/)
\end{abstract}
	
	}
\maketitle

\section{Introduction}
\label{sec:introduction}

\IEEEPARstart{B}{ased} on the excellent understanding and generation capabilities of Large Language Models (LLMs), Multimodal Large Language Models (MLLMs) are making progress in various tasks \cite{yang2025mm,lian2024gpt,zhan2026shiptraj}. 
With the widespread application of unmanned aerial vehicles (UAVs) globally and the rapid development of the low-altitude economy, many tasks currently need to be performed in the sky \cite{tian2025uavs,zhan2026mvpc}, such as UAV-based goods delivery, urban traffic/security patrol, industrial inspection, disaster rescue, and scenery tour \cite{feng2024security,zhan2023rsvg,zhao2025tcma}.
Since MLLMs can naturally combine UAVs' visual and text signals to complete various tasks more effectively, achieving this integration ability is crucial for advancing towards low-altitude intelligence.

However, the lack of reliable benchmarks to evaluate low-altitude UAV visual reasoning remains an obstacle.
In the past decade, deep learning models have achieved substantial success in various low-altitude UAV visual perception tasks, such as object detection \cite{zhu2021detection}, object tracking \cite{zhang2022webuav}, semantic segmentation \cite{lyu2020uavid}, and video classification \cite{mou2020era}. 
Recent efforts have explored limited vision-language tasks, including visual question answering (VQA) \cite{rahnemoonfar2021floodnet}, video captioning \cite{bashmal2023capera}, and referring expression comprehension \cite{sun2025refdrone} tasks. 
However, these benchmarks with a limited data scale focus only on a few specific tasks and do not adequately reflect real-world needs.

To fill the gap, we introduce \textbf{UAVBench}, a comprehensive and challenging vision-language benchmark for image-level and region-level understanding.
Following a common approach in the remote sensing domain \cite{hu2025rsgpt,yao2025falcon,zhan2025skyeyegpt,shabbirgeopixel}, a direct strategy is to collect and repurpose existing datasets, expanding upon them based on manual annotation or advanced large models.
The UAVBench comprises about 966k high-quality data samples and 43 test units, across 10 tasks at the image-level and region-level, covering 261k multi-spatial resolution and multi-scene images. We involve humans to ensure the high quality of the data.

We evaluate 11 state-of-the-art open-source and closed-source MLLMs on our UAVBench benchmark.
Our evaluation reveals the limited performance of the MLLMs on low-altitude tasks, and it is even unable to perform difficult tasks such as flight altitude estimation and object detection. 
Directly applying MLLMs to generate meaningful dialogue based on the visual content of low-altitude drones remains a challenge. 
MLLMs are usually trained on Internet data, which is quite different from low-altitude UAV data.
Low-altitude images often require complex scene understanding and spatial reasoning from multiple camera views, such as front-view, side-view, or bird-view.
As the flight altitude increases, the object becomes smaller (sometimes only 10 pixels).
MLLMs also need to respond normally to visual content under various weather conditions to meet the requirements of various tasks.
Based on our benchmark results, the practical demand urgently requires instruction-tuning data and specialized MLLMs for low-altitude UAVs.

\begin{figure*}
  \centering
  \includegraphics[width=0.99\linewidth]{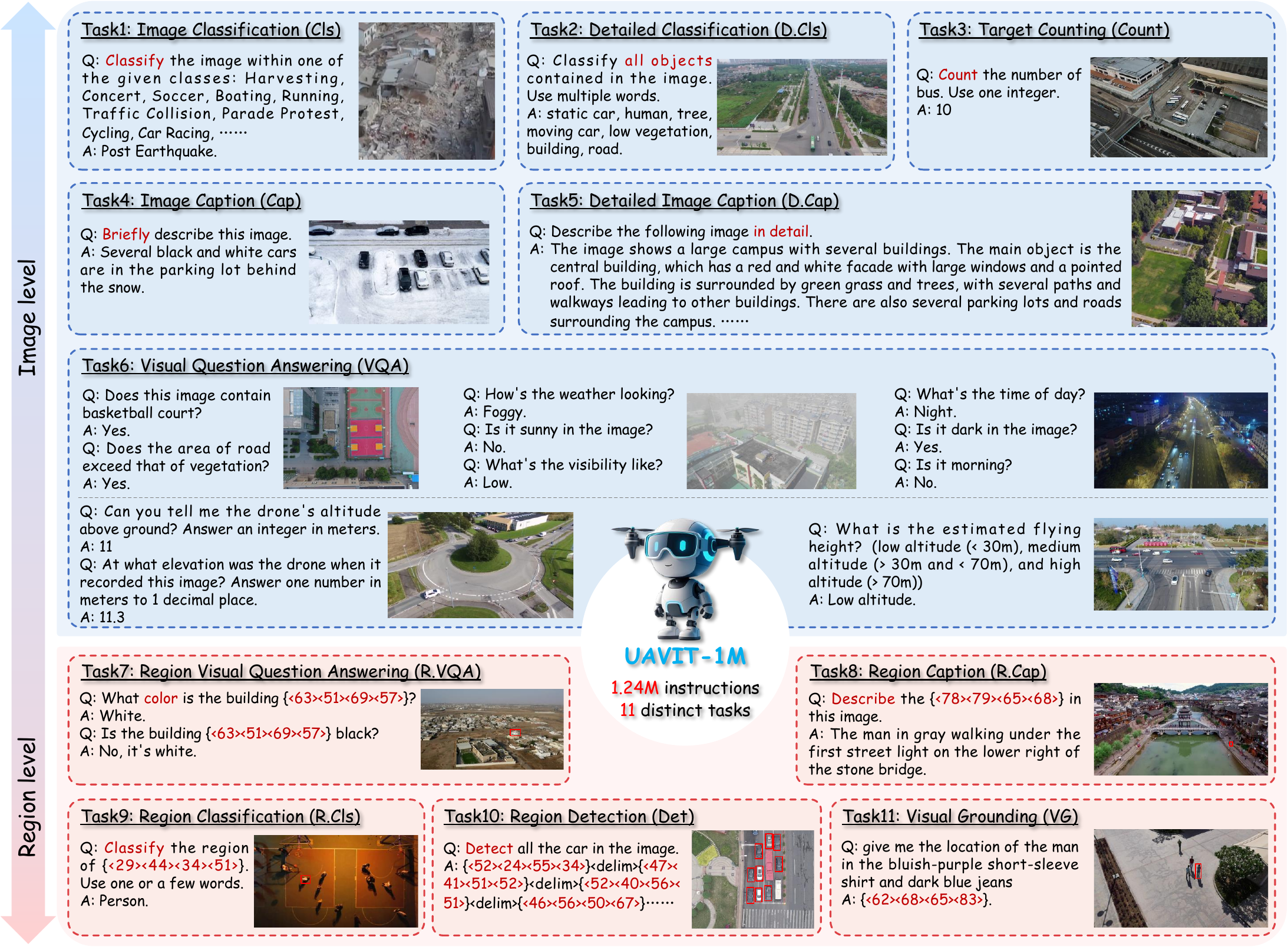}
  \caption{
  Sampled UAVIT-1M examples from our designed 11 tasks. We present the images, the corresponding questions, and the answers.
  These tasks are divided into two levels, namely, Image-level: Image Classification, Detailed Classification, Target Counting, Image Captioning, Detailed Image Captioning, and Image VQA; Region-level: Region VQA, Region Captioning, Region Classification, Region Detection, and Visual Grounding.
  }
  \label{fig:dataset}
\end{figure*}

To get one step closer to MLLMs for low-altitude UAVs, we further develop \textbf{UAVIT-1M}, a large-scale, multi-task instruction tuning dataset.
The image-text pair data are obtained using the same process used to build UAVBench, and then we organize the instruction tuning data by combining predefined templates and LLMs.
As shown in Figure \ref{fig:dataset}, UAVIT-1M consists of 1.24 million diverse instruction tuning conversations, 789k multi-scene low-altitude UAV images, and about 2,000 types of spatial resolution with 11 distinct tasks.
After fine-tuning open-source MLLMs with our UAVIT-1M, their performance on various tasks has been significantly improved.

Our main contribution can be summarized as follows:
\begin{itemize}
\item To the best of our knowledge, UAVBench is the first vision-language benchmark to assess low-altitude UAV image-level and region-level understanding and reasoning capabilities of MLLMs.
\item We conduct a comprehensive evaluation of open-source and closed-source MLLMs on UAVBench, revealing significant challenges in this field and providing valuable insights for future development.
\item UAVIT-1M stands as the first largest and most comprehensive instruction tuning dataset to enhance low-altitude visual understanding abilities, supporting 11 image-level and region-level tasks.
\item We conduct extensive experiments to demonstrate that fine-tuning MLLMs on UAVIT-1M leads to improved performance on UAVBench.

\item 
{We release the following assets to the public community for applications in real-world scenarios: 1) \href{https://huggingface.co/datasets/ZhanYang-nwpu/UAVBench}{{{\textit{UAVBench}}}} benchmark, 2) \href{https://huggingface.co/datasets/ZhanYang-nwpu/UAVIT-1M}{{{\textit{UAVIT-1M}}}} instruction tuning dataset, 
3) the \href{https://huggingface.co/ZhanYang-nwpu/LLaVA1.5-UAV}{{{\textit{LLaVA1.5-UAV}}}} model checkpoint, 
5) the \href{https://huggingface.co/ZhanYang-nwpu/MiniGPTv2-UAV}{{{\textit{MiniGPTv2-UAV}}}} model checkpoint, 
4) the \href{https://huggingface.co/ZhanYang-nwpu/GeoChat-UAV}{{{\textit{GeoChat-UAV}}}} model checkpoint.
}

\end{itemize}

We hope this effort will foster further research and development for the low-altitude UAV MLLMs, advancing the capabilities of MLLMs to solve low-altitude tasks and their real-world applications.

\section{Related Work}

{Inspired by Falcon\_SFT \cite{yao2025falcon}, a satellite remote sensing instruction-tuning dataset, this paper constructs a low-altitude UAV vision-language benchmark and instruction-tuning dataset for MLLMs. To contextualize our contributions, we summarize the related work on low-altitude visual datasets, low-altitude vision-language datasets, generic MLLMs, and remote sensing MLLMs (RS-MLLMs).
}

\subsection{Low-Altitude RGB Visual Datasets for UAVs}

The development of high-quality low-altitude visual perception datasets has attracted increasing attention in recent years. The single-modal RGB dataset mainly forms systematic benchmarks in areas such as object detection \cite{zhu2021detection}, semantic segmentation \cite{nigam2018ensemble, chen2018large, cai2025vdd}, object tracking \cite{zhang2022webuav}, and video classification \cite{perera2020multiviewpoint, mou2020era}.
{
Existing datasets generally fall into two categories: single-task and multi-task.
Single-task datasets target a single visual perception task. For instance, UAV123 \cite{mueller2016benchmark} and AU-AIR \cite{bozcan2020air} focus on tracking and detection, respectively, while DroneVehicle \cite{sun2022drone} and UAVid \cite{lyu2020uavid} are tailored for vehicle detection and semantic segmentation.
}
Multi-task datasets simultaneously support multiple visual perception tasks, such as UAVDT \cite{du2018unmanned}, VisDrone \cite{zhu2021detection}, and WebUAV-3M \cite{zhang2022webuav}.
In addition, adverse weather significantly increases the difficulty of low-altitude visual perception tasks, posing challenges to the model.
Datasets such as UAVDT, WebUAV-3M, and HazyDet \cite{feng2024hazydet} provide abundant data covering conditions such as rainy weather, foggy weather, night, low light, and motion blur.

\subsection{Low-Altitude Vision-Language Datasets for UAVs}
Despite the progress in visual perception, to the best of our knowledge, vision-language tasks for low-altitude UAVs are still in the exploratory stage.
The early FloodNet \cite{rahnemoonfar2021floodnet} includes the VQA dataset in flood scenarios, but the validation and test sets are not publicly available. 
{
While WebUAV-3M \cite{zhang2022webuav} is primarily an object tracking dataset, it also provides natural language descriptions for the bounding box of the first frame.
More recent contributions, including CapERA \cite{bashmal2023capera}, RefDrone \cite{sun2025refdrone}, and GeoText-1652 \cite{chu2024towards}, have curated datasets specifically for video captioning, referring expression comprehension, and natural language-guided geolocalization in drone views, respectively. 
Additionally, the novel task of spatial aerial video grounding has been facilitated by the introduction of the large-scale dataset UAV-SVG \cite{zhan2025where}.
Nevertheless, a comprehensive benchmark and instruction-tuning dataset covering a wide range of tasks remains absent.
}

\subsection{Generic Multimodal Large Language Models}
Current generic MLLMs have demonstrated substantial advancements in various vision and vision-language tasks. 
{
Pioneering works such as LLaVA \cite{liu2024improved} and MiniGPT-4 \cite{chen2023minigpt} have achieved notable success by using linear layers to connect LLMs and pretrained visual encoders.
Building on this foundation, subsequent research has focused on enhancing the spatial reasoning and grounding abilities of MLLMs.
}
Contributions from MiniGPTv2 \cite{chen2023minigpt}, Shikra \cite{chen2023shikra}, and Ferret-v2 \cite{zhang2024ferretv} further substantiate that LLMs can effectively process and detect spatial coordinates of objects within an image.

{
More recently, the field has witnessed a surge in models with enhanced architectural designs and scaling strategies. 
LLaVA-NeXT \cite{li2024llavanext} improves upon its predecessor by incorporating stronger language backbones and high-quality data mixtures to boost reasoning and OCR capabilities. 
Similarly, InternVL3 \cite{zhu2025internvl3} and InternVL2.5 \cite{chen2024internvl} explore advanced training recipes and scaling laws to align powerful vision foundation models with LLMs, achieving state-of-the-art performance on generic benchmarks. 
In the closed-source domain, models such as Claude 3 Sonnet \cite{Claude3} have exhibited exceptional visual reasoning speeds and accuracy, setting high standards for closed-source performance. 
Furthermore, models like InternLM-XC2 \cite{dong2024internlm} and Qwen2.5-VL \cite{bai2025qwen2} continue to advance general image understanding capacities. 
Concurrently, research attention is shifting towards fine-grained pixel-level reasoning. Models, including Osprey \cite{yuan2024osprey}, PixelLM \cite{ren2024pixellm}, PSALM \cite{zhang2024psalm}, and GLaMM \cite{rasheed2024glamm}, have achieved significant gains in pixel-level instruction following, enabling more precise segmentation and detailed visual understanding.
}

{
Recently, MM-InstructEval \cite{yang2025mm} focuses on general multimodal reasoning tasks with vision-text contexts (\textit{e.g.}, sentiment analysis, relation extraction) across 16 datasets and 6 task types.
MM-InstructEval does not include UAV-specific tasks, while UAVBench addresses this gap by integrating 21 existing open-source datasets.
Our UAVBench covers 10 image-level and region-level task types and provides 966k test samples with approximately 100 diverse instructions (vs. MM-InstructEval’s 34,602 test samples with 10 instruction templates), enabling rigorous evaluation of multimodal tasks for low-altitude scenarios. 
}

\subsection{Remote Sensing Multimodal Large Language Models}
Inspired by MLLMs in the natural image domain, remote sensing MLLMs have also made significant strides.
Based on open-source MLLMs, RSGPT \cite{hu2025rsgpt} and RS-LLaVA \cite{bazi2024rs} have made initial efforts in remote sensing single-granularity vision-language tasks.
Advanced works, including GeoChat \cite{kuckreja2024geochat} and SkyEyeGPT \cite{zhan2025skyeyegpt}, have introduced instruction-tuning techniques to equip MLLMs with multi-task instruction-following abilities effectively for RGB imagery.
EarthGPT-X and EarthGPT \cite{zhang2024earthgpt} expand visual input to include multisensory vision modalities such as RGB, infrared, and Synthetic Aperture Radar (SAR). 
{
Recent studies, such as GeoPixel \cite{shabbirgeopixel} and Falcon \cite{yao2025falcon}, have begun to focus intensely on improving pixel-level understanding and fine-grained reasoning within satellite imagery. 
However, these models are primarily trained on satellite data, leaving a significant performance gap when applied to the distinct viewpoints and object scales of low-altitude UAV scenarios.
}

\begin{figure*}[t]
  \centering
  \includegraphics[width=0.99\linewidth]{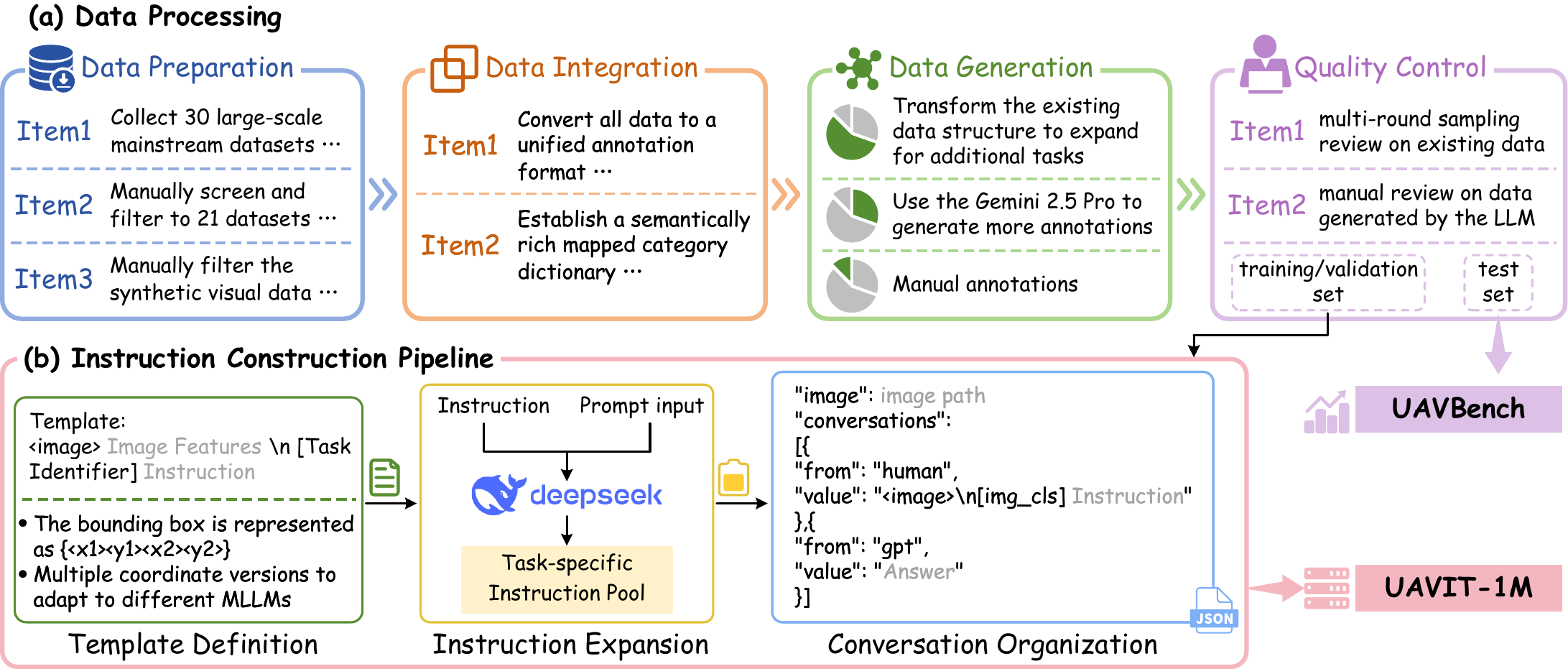}
  \caption{
  An overview of (a) data processing for UAVBench and (b) instruction construction pipeline for UAVIT-1M.
  }
  \label{fig:dataset_construct}
\end{figure*}

\section{UAVBench Benchmark and UAVIT-1M Dataset}
Although many MLLMs now exhibit good performance in various general tasks, they are difficult to meet the practical application requirements of low-altitude UAV scenarios. Under a low-altitude bird's-eye view, the image understanding and target recognition capabilities of MLLMs deteriorate notably.
To evaluate the low-altitude visual dialogue capabilities of existing models, we develop the first large-scale UAVBench benchmark.
To equip MLLMs with powerful capabilities in low-altitude image-level and region-level perception and reasoning, we further introduce UAVIT-1M, the first multi-task low-altitude visual instruction-tuning dataset.

\begin{table*}[t]
\addtolength{\tabcolsep}{-4pt}
\centering
\caption{
{
The 21 selected low-altitude visual or vision-language datasets for the construction of UAVBench and UAVIT-1M.
Supported tasks include Semantic Segmentation (SS), Object Detection (OD), Video Captioning (VC), Event Recognition (ER), Natural Language-guided Geolocalization (NLG), Action Recognition (AR), Referring Expression Comprehension (REC), Object Tracking (OT), Multiple Object Tracking (MOT), and Image-Text Retrieval (ITR). $\dag$ indicates that  there might be visual data in this source that does not originate from the real world.
}
}
\label{tab:all_datasets_list}
\resizebox{1\linewidth}{!}{
\begin{tabular}{cccccccc}
\toprule
\multirow{2}{*}{Dataset}   & Official   & Dataset   & Number of Images in & \multirow{2}{*}{Spatial Resolution}  & \multirow{2}{*}{\#Class}  & Supported & \multirow{2}{*}{Visual Data Sources} \\
 & Link & Size & UAVIT-1M (UAVBench) &   & & Tasks &  \\
\midrule
AeroScapes \cite{nigam2018ensemble}   & \href{https://github.com/ishann/aeroscapes}{Link}  & 752M & 2,621 (648) & 1280$\times$720   & 12 & SS & commercial drone \\
AU-AIR \cite{bozcan2020air}   & \href{https://bozcani.github.io/auairdataset}{Link}   & 2.2GB   & 30,000 (2,823)  & 1920$\times$1080   & 8     & OD   & quadrotor (Parrot Bebop 2) \\
CapERA \cite{bashmal2023capera}   & \href{https://github.com/yakoubbazi/CapEra/tree/main/dataset}{Link}  & 1.46GB   &   2,568 (296) & 640$\times$640   &  25  &  VC & ERA dataset\\
DroneVehicle \cite{sun2022drone}   & \href{https://github.com/VisDrone/DroneVehicle?tab=readme-ov-file\#dataset}{Link}  & 13.05GB   & 18,920 (8,734) & 840$\times$712   &   5 & OD & DJI M200 aircraft \\
ERA \cite{mou2020era}   & \href{https://lcmou.github.io/ERA\_Dataset/}{Link}  & 6.29GB   & 22,521 (2,607)  & 640$\times$640   &  25  & ER & from Youtube {$\dag$}\\  
FloodNet  \cite{rahnemoonfar2021floodnet}   & \href{https://www.dropbox.com/scl/fo/k33qdif15ns2qv2jdxvhx/ANGaa8iPRhvlrvcKXjnmNRc?rlkey=ao2493wzl1cltonowjdbrnp7f}{Link}   & 12.1GB   & 1,445 (0)  & 4000$\times$3000, 4592$\times$3072   & 2  & SS \& VQA & DJI Mavic Pro quadcopter   \\ 
GeoText-1652 \cite{chu2024towards}   & \href{https://github.com/MultimodalGeo/GeoText-1652/blob/main/README.md\#-download-links}{Link}   & 6.2GB   & 37,854 (0)  & 512$\times$512   & 701  & NLG  &  drone platform\\
HazyDet \cite{feng2024hazydet}   & \href{https://github.com/GrokCV/HazyDet?tab=readme-ov-file\#hazydet}{Link}  & 8.41GB   & 18,000 (4,000)  & $480\times360 \sim 4096\times2250$   &    3& OD & drone platform\\
MOD20 \cite{perera2020multiviewpoint}   & \href{https://asankagp.github.io/mod20/}{Link}  & 10.6GB   & 0 (15,313)  & 720$\times$720   &  20  &  AR & from Youtube {$\dag$}\\ 
RefDrone \cite{sun2025refdrone}   & \href{https://github.com/sunzc-sunny/refdrone}{Link}  & 0.28GB   & 0 (1,503)  & $960\times540 \sim 1920\times1080$   &  10  & REC & VisDrone2019 dataset\\
RDDTS \cite{feng2024hazydet}   & \href{https://github.com/GrokCV/HazyDet?tab=readme-ov-file\#hazydet}{Link}  & 0.06GB   &  0 (600) & $300\times129 \sim 2950\times2094$   & 3   & OD & drone platform\\
Semantic Drone \cite{SemanticDrone}   & \href{http://dronedataset.icg.tugraz.at}{Link}  & 3.9GB   & 400 (0)& 6000$\times$4000   & 22 &SS  & drone platform\\
UAV123 \cite{mueller2016benchmark}   & \href{https://drive.usercontent.google.com/download?id=0B6sQMCU1i4NbNGxWQzRVak5yLWs}{Link}  & 13.0GB   & 109,895 (0)& 720$\times$480, 1280$\times$720   & 10 & OT & \multicolumn{1}{c}{\begin{tabular}[c]{@{}c@{}}UAV (DJI S1000)\\ small low-cost UAV \\ UAV simulator {$\dag$} \end{tabular}}  \\
UAVDT-M \cite{du2018unmanned}   & \href{https://sites.google.com/view/grli-uavdt/}{Link}  & 6.32GB   &  24,143 (16,592) & 960$\times$540, 1024$\times$540   &  3  & MOT \& OD &  UAV platform \\
UAVDT-S \cite{du2018unmanned}   & \href{https://sites.google.com/view/grli-uavdt/}{Link}  & 4.83GB   & 0 (37,084)  & 960$\times$540, 1024$\times$540  &  3  & OT & UAV platform\\  
UAVid \cite{lyu2020uavid}   & \href{https://uavid.nl/}{Link}  & 6.18GB   & 270 (0) & 3840$\times$2160, 4096$\times$2160    &  8  &SS& drones (DJI-Phantom 3 \& 4)  \\
UDD \cite{chen2018large}   & \href{https://github.com/MarcWong/UDD?tab=readme-ov-file\#download-links}{Link}  & 1.1GB   & 106 (35) & 3840$\times$2160, 4096$\times$2160, 4000$\times$3000   & 6  &SS & UAV (DJI-Phantom 4) \\
UDV DIT \cite{huang2024visual}   & \href{https://github.com/huangjh98/VCSR?tab=readme-ov-file\#dataset}{Link}   & 2.79GB   & 8,393 (2,102)  & $110\times83 \sim 5464\times6000$  & $/$    & ITR   & drone platform\\
VDD \cite{cai2025vdd}  & \href{https://github.com/RussRobin/VDD?tab=readme-ov-file\#download-links}{Link}  & 2.05GB   & 360 (40) & 4000$\times$3000   & 7 &SS  & DJI Mavic AIR II \\
VisDrone2019-DET \cite{zhu2021detection}   & \href{https://github.com/VisDrone/VisDrone-Dataset?tab=readme-ov-file\#download}{Link}  & 1.79GB   &  7,019 (1,609) & $480\times360 \sim 2000\times1500$   & 12 &  OD & DJI Mavic, Phantom series \\
WebUAV-3M \cite{zhang2022webuav}   & \href{https://github.com/983632847/WebUAV-3M?tab=readme-ov-file\#dataset-download}{Link}  & 819GB   & 165,282 (11,879) & $432\times320 \sim 2160\times2018$   & 223  & OT & from Youtube {$\dag$}\\ 
\bottomrule
\end{tabular}
}
\end{table*}

\subsection{Data Processing and Benchmark Construction}
No existing dataset adequately meets the specialized demands of evaluating MLLMs for low-altitude vision-language perception.
Following Falcon\_SFT \cite{yao2025falcon}, we employ a methodical yet straightforward method that involves integrating and repurposing a wide array of open-source datasets to construct the low-altitude vision-language benchmark.
Our pipeline is organized into the following four stages, as illustrated in Figure \ref{fig:dataset_construct}.

\subsubsection{Step 1: Data preparation}
Our process begins with an extensive data collection phase.
We gather 30 large-scale mainstream datasets that are prominent in the UAV visual perception community, such as VisDrone, WebUAV-3M, and AnimalDrone.
This initial database undergoes a meticulous manual review process, resulting in the selection of 21 publicly available datasets deemed most suitable for our objectives.
To facilitate reproducibility and further research, we have compiled a comprehensive list of these 21 datasets, complete with their metadata and direct download links, as detailed in Table \ref{tab:all_datasets_list}. We accept only visual data from the real world.
{
Subsequently, for UAV123 \cite{mueller2016benchmark} with synthetic visual data, we follow the explicit synthetic labels to delete the synthetic sequences rendered by the Unreal4 Engine.
For datasets crawled from web platforms (marked with † in Table \ref{tab:all_datasets_list}), including 9,688 videos from ERA, MOD20, and WebUAV-3M, we conduct a comprehensive manual review to identify potential synthetic content. We confirm that all these videos are captured in the real world.
}

\begin{table*}[t]
\centering
\caption{Semantically rich mapped category dictionary. The object categories are all from UAVBench and UAVIT-1M. The mapping ID is manually annotated.
 }
\label{tab:Mapped category}
\resizebox{1\linewidth}{!}{
\begin{tabular}{ccc||ccc||ccc||ccc}
\toprule
\textbf{ID} & \textbf{Category} & \textbf{Mapped} & \textbf{ID} & \textbf{Category} & \textbf{Mapped} & \textbf{ID} & \textbf{Category} & \textbf{Mapped} & \textbf{ID} & \textbf{Category} & \textbf{Mapped} \\
\midrule
1 & agricultural machinery & & 2 & aircraft &  & 3 & animal & & 4 & ar-marker & \\
5 & artifact & &6 & awning-tricycle &89/94 &7 & backpacking & &8 & bald tree & 65/88/93 \\
9 & baseball & &10 & basketball & &11 & bicycle & 12 &12 & bike & 11 \\
13 & bird & 3 &14 & boat & 95&15 & boating & &16 & building & 25 \\
17 & bus &70/94 &18 & car & 94&19 & car racing & &20 & chainsawing trees &  \\
21 & cliff jumping & &22 & concert & &23 & conflict & &24 & constructing &  \\
25 & construction & 16&26 & cutting wood & &27 & cycling & &28 & dancing &  \\
29 & dirt & &30 & dog &3 &31 & door & &32 & drone & 2/91 \\
33 & fence & &34 & fence pole & &35 & fighting & &36 & figure skating &  \\
37 & fire & &38 & fire fighting & 37 &39 & flood &98 &40 & freight car & 94 \\
41 & grass & 50/65/93&42 & gravel & &43 & group & 45/63/64 &44 & harvesting &  \\
45 & human &63/64 &46 & industry machine & &47 & jetskiing & &48 & kayaking &  \\
49 & landslide & &50 & low vegetation & 65/93&51 & motor & 52&52 & motorbike & 51 \\
53 & motorbiking & &54 & moving car & 18/94&55 & mudslide & &56 & national football&  \\
57 & natural object & &58 & obstacle & &59 & parade protest & &60 & party &  \\
61 & paved area & &62 & pedestrian &45/63/64 &63 & people &45/64 &64 & person & 45/63 \\
65 & plant &93 &66 & ploughing & &67 & police chase & &68 & pool & 98 \\
69 & post earthquake & &70 & public transport & &71 & religious activity & &72 & road &  \\
73 & rock climbing & &74 & rocks & &75 & roof & &76 & running &  \\
77 & skateboarding & &78 & skiing & &79 & sky & &80 & soccer &  \\
81 & standup paddling & &82 & static car &18/94 &83 & surfing & &84 & swimming &  \\
85 & traffic collision & &86 & traffic congestion & &87 & trailer & 94 &88 & tree & 65/93\\
89 & tricycle & 94&90 & truck &94 &91 & uav & 2/32&92 & van & 94 \\
93 & vegetation &65 &94 & vehicle & &95 & vessel & 14&96 & wakeboard &  \\
97 & wall & &98 & water body & &99 & window & &100 & windsurfing &  \\

\bottomrule
\end{tabular}
}
\end{table*}

\subsubsection{Step 2: Data Integration}
Following the preparation phase, the subsequent challenge is to integrate the 21 disparate datasets into a cohesive whole. During this process, two primary sources of heterogeneity must be handled.
The first is the variation in annotation formats across different selected datasets, such as the YOLO format versus the COCO format for the bounding boxes. We resolve this by defining a standardized annotation schema and converting all datasets to this unified format.
The second and more complex issue is the inconsistency in category nomenclature across different datasets.
a) A single object type might be labeled with different terms. For instance, "people" in the VisDrone dataset \cite{zhu2021detection} is labeled as "person" in the UAV123 dataset \cite{mueller2016benchmark} and as "human" in the AU-AIR dataset \cite{bozcan2020air}.
b) Furthermore, there are overlaps in the semantic scope, where a label like "bus" in some datasets \cite{du2018unmanned, feng2024hazydet} could fall under a broader category like "public transport" in another \cite{zhang2022webuav}. 

To overcome the above issues, we manually annotate a comprehensive and semantically rich mapped category dictionary, as shown in Table \ref{tab:Mapped category}. 
In the real world, the same object usually has different names. We map each category name to a label and simultaneously map it to other category names that can represent the same objects. Random mapping is performed during model training.
In Falcon\_SFT \cite{yao2025falcon}, only a unique and uniform category name is reserved for each object. Although this ensures consistent category naming across datasets, it introduces bias and undermines the semantic generalization of the model. 
Our mapping category dictionary unifies category naming and also enriches semantic expression. When the model recognizes "car", "bus", "truck", and "van" as "vehicle", we also consider it correct.

\begin{table*}[t]
\centering
\caption{ The specific explanation of data generation for the 11 new low-altitude vision-language tasks shown in Figure \ref{fig:dataset}.
}
\label{tab:taskto11}
\resizebox{1\linewidth}{!}{
\begin{tabular}{lll}
    \toprule
    Raw Task & New Task & Data Generation Method  \\
    \midrule
    Recognition & Image Classification & Set the category of the image as the answer. \\
    SS & Detailed Classification & Set the category of all objects contained in the image as the answer. \\
    OD & Counting Target & Set a category as the question and the total number of the corresponding boxes as the answer. \\
    VC & Image Caption & Given the raw description sentence, use the LLM to create captions as the answer. \\
    ITR & Image Caption & Set the raw description sentence as the answer. \\
    NLG & Detailed Image Caption & Set a description paragraph as the answer. \\
    Image VQA & Image VQA & Use the original question-answer pairs. No changes. \\
    SS & Image VQA (Presence)&  Given all objects contained in the image, use the LLM to create "Presence" questions. \\
    SS & Image VQA (Comparison)& Based on the pixel areas of each object in the segmentation mask, use the LLM to create "Comparison" questions\\
    OD and OT & Image VQA (Weather)& Given weather conditions or lighting conditions of the image, use the LLM to create "Weather" questions.  \\
    OD and OT & Image VQA (Altitude)& Given the flight altitude category or precise altitude of the image, use the LLM to create "Altitude" questions.  \\
    Visual Grounding & Region VQA & Given the referring expression of the bounding box, use the LLM to create the question and answer of the object. \\ 
    OT & Region Caption & Set a bounding box as the question, and the manually annotated description as the answer. \\
    Visual Grounding & Region Caption & Set a bounding box as the question, and the corresponding referring expression as the answer. \\
    OD and OT & Region Classification & Set a bounding box as the question, and set the corresponding category as the answer. \\
    OD and OT & Region Detection & Set a category as the question and set all the corresponding bounding boxes as the answer. \\
    REC & Visual Grounding & Set the referring expression as the question, and the corresponding bounding box as the answer. \\
    OT & Visual Grounding & Set a manually annotated description as the question, and the corresponding bounding box as the answer. \\
    \bottomrule
    \end{tabular}
    }
\end{table*}

 \begin{table*}[t]
 \renewcommand\tabcolsep{2pt} 
 \centering
  \caption{Image and annotation statistics on UAVBench and UAVIT-1M.}
\label{tab:statistics}
 \resizebox{0.99\linewidth}{!}{
 \begin{tabular}{clccc}
 \toprule
 \textbf{Spatial hierarchy} & \textbf{Tasks} & \textbf{\#images}& \textbf{\#Annotations} & \textbf{Data sources} \\
 \midrule
\textbf{UAVBench} & &\textbf{261k} &\textbf{966k} & \\
 \multirow{5}{*}{\rotatebox[origin=c]{0}{\centering Image Level}}
&~- Cls & 17,920  & 17,920  & ERA, MOD20\\
&~- D.Cls & 723 & 723  & AeroScapes, VDD, UDD\\
&~- Cap & 2,398 & 11,990  & CapERA, UDV DIT\\
&~- VQA & 51,609&  52,912 & VDD, UDD, AeroScapes, HazyDet, UAVDT-M, AU-AIR, UAVDT-S\\
&~- Count & 34,181  & 62,537 & AU-AIR, DroneVehicle, HazyDet, VisDrone2019-DET, RDDTS, UAVDT-M\\ 
\cmidrule(lr){1-1}
\multirow{5}{*}{\rotatebox[origin=c]{0}{\centering Region Level}}
&~- R.Cls & 83,036  & 732,197  &AU-AIR, WebUAV-3M, RefDrone, DroneVehicle, HazyDet, RDDTS, UAVDT-M, UAVDT-S\\
&~- R.Cap & 18,430 &18,430 &WebUAV-3M, UAVDT-S\\
&~- R.VQA & 18,698 &18,698  &UAVDT-S\\
&~- Det & 26,516 & 43,208 &AU-AIR, DroneVehicle, HazyDet, UAVDT-M, VisDrone2019-DET, RefDrone, RDDTS\\
&~- VG & 7,699 &7,699 & RefDrone, UAVDT-S\\
  \midrule
\textbf{UAVIT-1M} && \textbf{789k} & \textbf{1.24M} & \\
 \multirow{6}{*}{\rotatebox[origin=c]{0}{\centering Image Level}}
&~- Cls & 22,521 &22,521 & ERA\\
&~- D.Cls & 3,757 & 3,757  & AeroScapes, Semantic Drone, VDD, UDD, UAVid \\
&~- D.Cap & 37,854 & 113,562 & GeoText-1652\\
&~- Cap & 10,961 & 54,805 &CapERA, UDV DIT\\
&~- VQA &72,449 & 172,838  &FloodNet, VDD, UDD, UAVid, AeroScapes, HazyDet, UAVDT-M, AU-AIR\\
&~- Count & 73,939 &156,241 & AU-AIR, DroneVehicle, HazyDet, VisDrone2019-DET\\
\cmidrule(lr){1-1}
\multirow{5}{*}{\rotatebox[origin=c]{0}{\centering Region Level}}
&~- R.Cls &181,873  &272,463 & AU-AIR, WebUAV-3M, UAV123\\
&~- R.Cap & 116,995  & 116,995& WebUAV-3M\\
&~- R.VQA &  95,203  & 95,203  &UAV123\\
&~- Det & 81,557 &140,207 & AU-AIR, DroneVehicle, HazyDet, UAVDT-M, VisDrone2019-DET\\
&~- VG & 92,074  & 92,074  &UAV123\\
 \bottomrule
 \end{tabular}
 }
 \end{table*}

 \begin{figure*}[t]
 \centering
\includegraphics[width=0.99\linewidth]{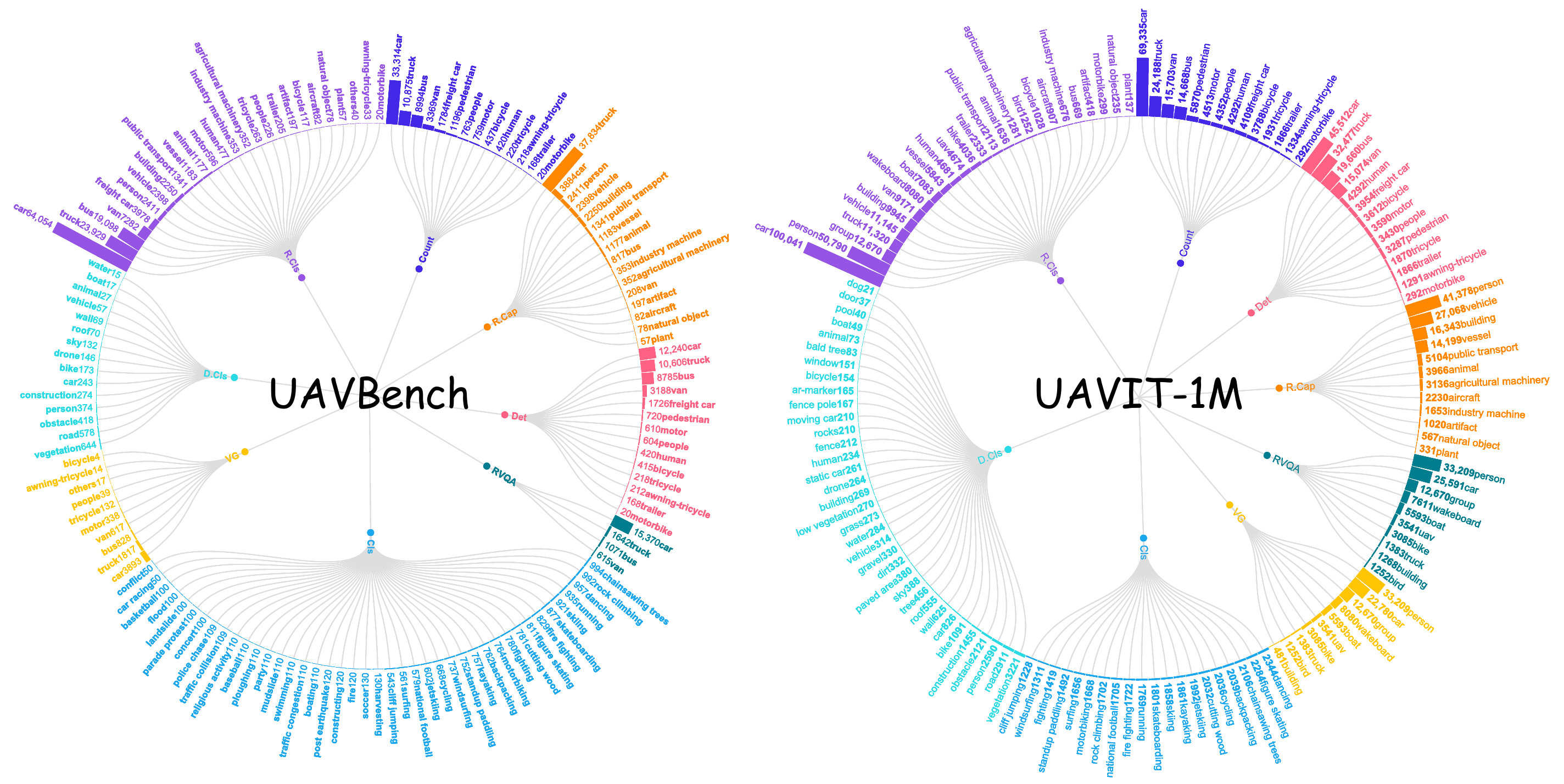}
 \caption{
 Category distribution in each task. Zoom in to view the specific categories and corresponding quantities.
 }
 \label{fig:object_distribution}
 \end{figure*}

\begin{figure*}[t]
\centering
\includegraphics[width=\textwidth]{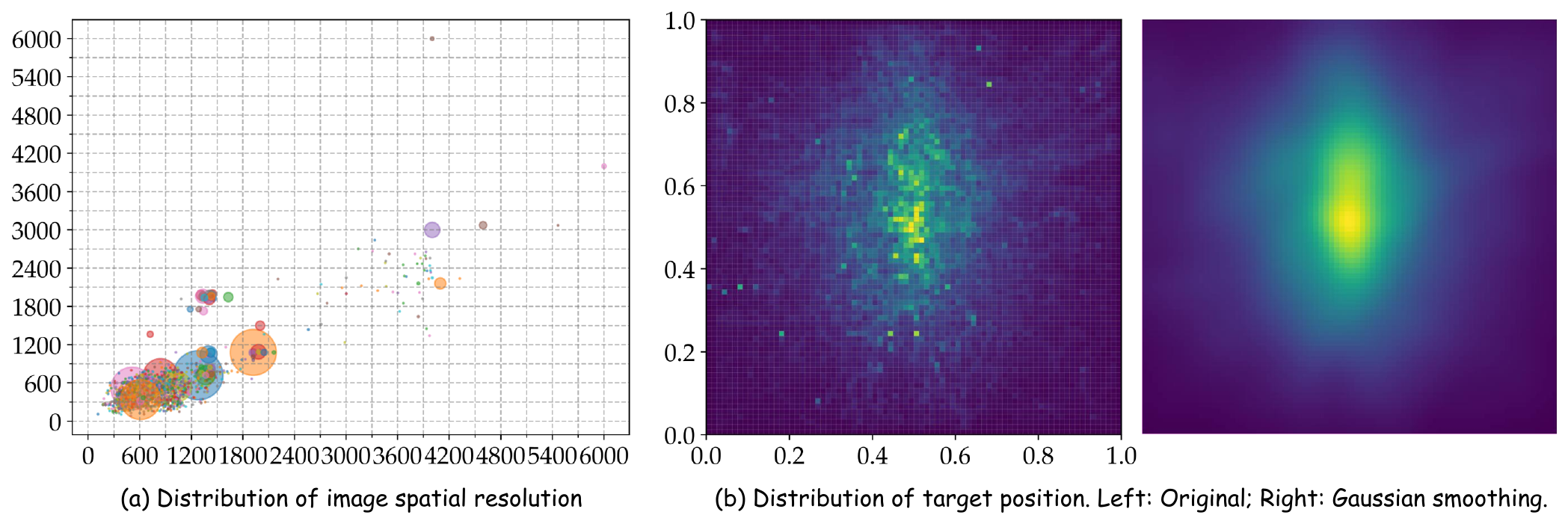}
\caption{Image resolution and target position distributions. Best viewed by zooming in.}
\label{fig:dataana2}
\end{figure*}

\subsubsection{Step 3: Data Generation}
The raw tasks of the selected dataset are primarily rooted in object detection and tracking, semantic segmentation, and recognition, complemented by five multimodal tasks with five datasets (VC, VQA, NLG, REC, and ITR tasks in Table \ref{tab:all_datasets_list}).
To enhance the utility of our dataset for advanced MLLMs and to cover a broader application of the real world, we undertake a data generation phase. Under significant resource limitations, we expand the dataset through three targeted methods, as shown in Figure \ref{fig:dataset_construct} (a).
This generation phase includes simple data structure transformation, generation based on Gemini 2.5 Pro \cite{comanici2025gemini}, and manual annotation. 
{The proportions of data in each category are 79.7\%, 11.1\%, and 9.2\% in sequence.}
The specific explanation of data generation for the 11 new low-altitude vision-language tasks is presented in Table \ref{tab:taskto11}.
{
During the manual annotation stage, we employ a double-check mechanism where each processed data is cross-reviewed by a second expert to ensure consistency across different annotation experts. His task is to evaluate whether the expressions annotated by others are correct. In cases of disagreement, a third senior annotator will act as the final checker to provide the final arbitration.
}

\subsubsection{Step 4: Quality Control}
A crucial final step is implementing a stringent review measure to ensure the quality of our benchmark and instruction-tuning data.
{
For the LLM-generated data, this part includes image captioning, image VQA, and region VQA tasks. To avoid errors or hallucinations, we conduct a full review of the data generated by the LLM for image captioning and region VQA tasks. Considering that image VQA uses LLM and given real labels to create questions, no intensive labor is spent on complete verification. Specifically, 115.7k samples are manually reviewed. The percentage of samples corrected manually is 0.39\%.
For the portion involving existing dataset expansion, we perform two rounds of random manual quality scoring. Specifically, 0.6\% of the samples are randomly sampled from each task in each round, and approximately 20k samples are reviewed in this stage. Finally, the rejection rate by the checker is a mere 0.05\%. This extremely low rate for the final pass demonstrates the effectiveness of our data construction in achieving non-ambiguous and high-quality annotations.
}
To prevent label leakage, we only consider the test set of the source dataset to construct the UAVBench benchmark and the training/validation set to construct the instruction fine-tuning data.

\subsection{Instruction-Tuning Dataset Construction}
While instruction-tuning datasets are readily available in the general domain, there are no sufficient resources for low-altitude UAV scenarios. As shown in Figure \ref{fig:dataset_construct} (b), we systematically convert our integrated image-text pairs into UAVIT-1M, a comprehensive, large-scale, real-world multimodal instruction-tuning dataset for low-altitude MLLM training.
\textbf{Step1: Template Definition.}
Drawing inspiration from established practices in the field \cite{liu2024improved, kuckreja2024geochat, yao2025falcon}, we first design unified instruction templates tailored to each specific low-altitude task, \textit{i.e.}, $<$image$>$ Image Features \textbackslash n[Task Identifier] Instruction.
We incorporate task-specific identifiers, such as "[img\_cap], [reg\_vqa], [vg]", prepended to each instruction.
This approach not only unifies the diverse low-altitude tasks but also empowers the model to generate task-specific outputs with greater flexibility.
Furthermore, the bounding boxes involved in the specific input instructions are represented as \{$<x_1><y_1><x_2><y_2>$\}, indicating the coordinates of the top left and bottom right vertices. For the Region VQA task, "What color is the building \{$<63><51><69><57>$\}?" is given.
To adapt to a variety of MLLMs, we offer multiple coordinate versions: an absolute coordinate, one scaled to hundredths, and another to thousandths. 
\textbf{Step2: Instruction Expansion.}
To enhance the model's ability to understand diverse instructions and to prevent it from becoming dependent on specific expressions, we use DeepSeek \cite{wu2024deepseek} to enrich the instruction pool for each task. For example, "Count the number of buses." can be expanded to "How many buses are there?", "Provide the quantity of buses.", \textit{etc}. 
\textbf{Step3: Conversation Organization.}
In the final stage, we assemble the conversation instruction-tuning data. For each image, we organize task-specific inputs using predefined templates and the extended instruction pool. Subsequently, these inputs are paired with their corresponding annotations to create complete conversation instances.

\begin{figure*}[t]
  \centering
  \includegraphics[width=0.99\linewidth]{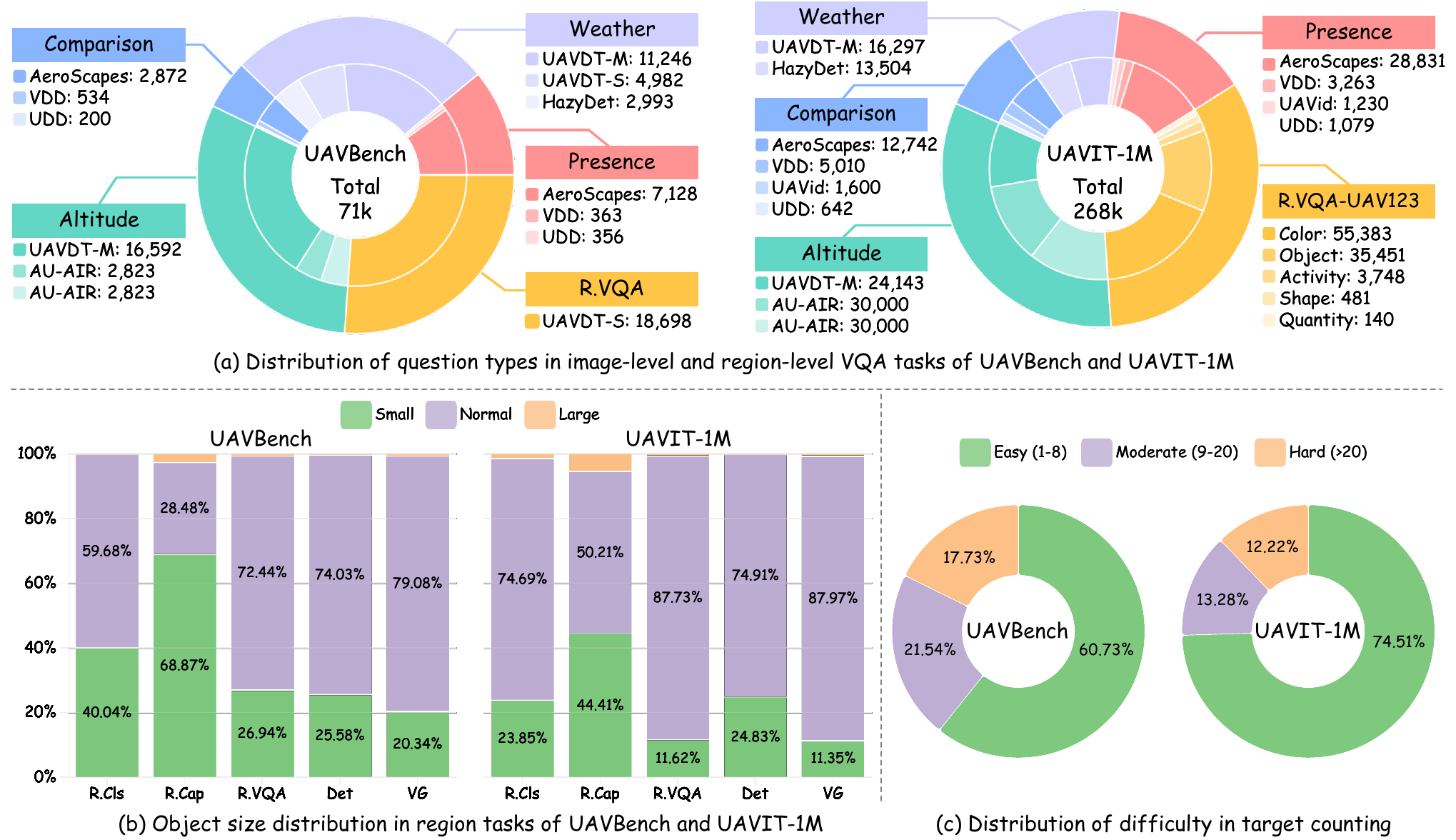}
  \caption{
  Some typical statistics of UAVBench and UAVIT-1M. (a) Distribution of question types in image and region-level VQA tasks. (b)  Distribution of object sizes in all region-level tasks. (c) Distribution of difficulty in the target counting task.
  }
  \label{fig:dataana}
\end{figure*}

\subsection{Dataset Statistics}
The data statistics and detailed data sources for each task are provided in Table \ref{tab:statistics}.
In UAVBench, we built 966k test samples across 261k images and 10 tasks, with a total of 43 test units (\textit{i.e.}, 43 test columns of Tables \ref{tab:imageclsVQA} - \ref{tab:regioncap}).
The UAVIT-1M dataset features a total of 1.2M instruction-tuning data and 789k images covering about 2,000 types of spatial resolution, supporting 11 image-level and region-level tasks.

\textbf{Object Category.}
Our UAVBench and UAVIT-1M contain various scenarios in the real world, such as rural areas, cities, parking lots, residential areas, universities, disaster-stricken areas, streets, transportation hubs, multiple weather and lighting condition scenarios, \textit{etc}., covering a wide range of object categories. 
Figure \ref{fig:object_distribution} shows the category distribution of each task. All 100 object categories are shown in Table \ref{tab:Mapped category}.

\textbf{Spatial Resolution of Image.}
In Figure \ref{fig:dataana2} (a), the distribution pattern of the image resolution in UAVIT-1M is illustrated by the different color circle distributions and the area of circles. Our resolution is widely distributed and contains 1,993 types.
The (minimum, maximum) values of the width and height are (110, 6000) and (83, 6000), respectively.

\textbf{Position Distribution of Objects.}
{
We confirm through distribution that objects are not randomly scattered or artificially clustered (for example, due to low-quality annotations).}
The distribution of the normalized target center positions in UAVIT-1M is shown in Figure \ref{fig:dataana2} (b).
{
The research objects of our dataset are mainly distributed near the central axis of $x=0.5$. The position distribution presents a central mean Gaussian and is a realistic reflection of real-world UAV operation scenarios, where objects of interest are intentionally kept near the center of the field of view to ensure clarity and relevance for downstream tasks.
This central distribution does not directly reflect specific viewing angles or flight patterns. It only indicates that the 2D pixel coordinates of research objects in regional-level tasks are mainly distributed around the center of the images. 
}

\textbf{Object Size Distribution.}
The object size distribution of region-level tasks is demonstrated in Figure \ref{fig:dataana} (b).
Due to the diverse image resolutions, the absolute area of the bounding box cannot accurately measure the object size. 
We define the area of the bounding box divided by the image area, and then take the square root to measure the object size. 
The object size is small ($<$3\%), normal ($\geq$ 3\% and $\leq$ 30\%), or large ($>$30\%).
UAVBench and UAVIT-1M have similar distributions. However, UAVBench faces more serious small-target challenges, especially for R.Cap and R.Cls tasks.

\textbf{Question Types of VQA Task.}
We show the statistics of question types in Figure \ref{fig:dataana} (a). We provide 71k and 268k question-answer pairs in UAVBench and UAVIT-1M.  
Image-level questions include presence, comparison, weather condition recognition, flight altitude recognition, and flight altitude estimation (integer and decimal), as shown in Table \ref{tab:imageclsVQA}.
Region-level questions include color, object, activity, shape, \textit{etc}.

\textbf{Difficulty of Target Counting.}
The target counting task based on UAVs is of great significance to tasks such as traffic management, crowd counting, and urban planning.
We show the counting difficulty distribution in Figure \ref{fig:dataana} (c).
According to the complexity of human counting of different numbers of objects, we classify the problems into easy (1-8), moderate (9-20), and hard ($>$20). It shows that the UAVBench faces more moderate and hard counting challenges.

{
\textbf{Task Distribution.}
The sample size of each task in our UAVIT-1M dataset is shown in Table \ref{tab:statistics}. The tasks are inherently imbalanced due to the varying scales of the 21 source datasets. Among the 11 tasks, R.Cls has the largest scale, accounting for 21.9\%, while D.Cls has the smallest scale, accounting for 0.4\%.
}

\begin{table*}[]
\centering
\caption{
{Details of the baseline model structure and weight fine-tuning ($\lozenge$: frozen weights; $\blacklozenge$: fine-tuning with LoRA techniques).}
}
\label{tab:modelset}
\begin{tabular}{lccc}
\toprule
\textbf{Model} & \textbf{Vision Modality Encoder} & \textbf{V-L  Alignment Layer} & \textbf{LLM Backbone} \\
\midrule
MiniGPTv2-UAV  & Eva-CLIP ViT @448 $^\lozenge$ & Linear Projector $^\blacklozenge$  & LLaMA-2-Chat-7B $^\blacklozenge$ \\
LLaVA1.5-UAV  & CLIP ViT-L/14 @336 $^\lozenge$ & MLP $^\blacklozenge$  & Vicuna-v1.5-7B $^\blacklozenge$  \\
GeoChat-UAV  & CLIP ViT-L/14 @336 $^\lozenge$ & MLP $^\lozenge$  & Vicuna-v1.5-7B $^\blacklozenge$ \\
    \bottomrule
\end{tabular}
\end{table*}

\begin{figure}[]
\centering
\includegraphics[width=0.99\linewidth]{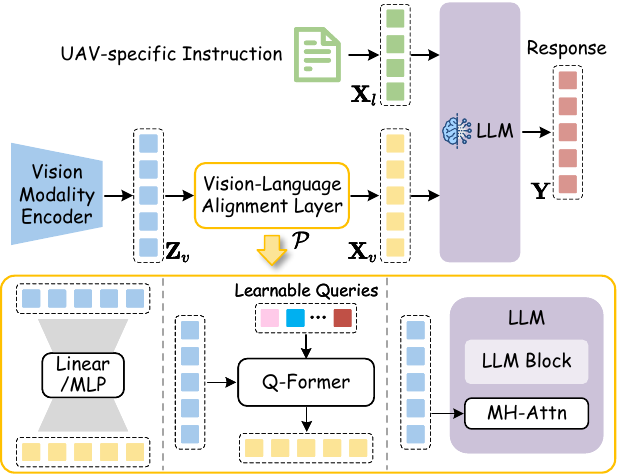}
\caption{
{Typical architecture of Multimodal Large Language Models (MLLMs), composed of a vision modality encoder, a vision-language alignment layer, and a large language model. Three types of alignment layers are mainly used, namely the linear layer or MLP, the Q-Former model, and the additional cross-attention layer. 
}
}
\label{fig:VLalign}
\end{figure}

\section{Multimodal Fusion and Alignment Methodology}
{
MLLMs bridge the gap between modalities by utilizing a specific vision-language alignment layer to connect a visual encoder, which extracts high-level image features, and a large language model, which enables text-based conversational reasoning. A diagram of the architecture is shown in Figure \ref{fig:VLalign}.
To address the multimodal alignment required in low-altitude UAV scenarios, we continue this alignment strategy, which serves as the modality interface $\mathcal{P}$.
In the existing MLLMs, three types of alignment layers are mainly used, namely the linear layer or MLP, the Q-Former model, and the additional cross-attention layer. Linear projector is an elementary method and also the one with the best performance and the widest application.
}

{
Given a low-altitude image $I \in \mathbb{R}^{H \times W \times C}$, the visual encoder extracts high-level latent features $\mathbf{Z}_v \in \mathbb{R}^{N \times D_v}$, where $N$ is the number of visual tokens and $D_v$ is the visual embedding dimension. To bridge the modality gap, the alignment layer $\mathcal{P}$ transforms these features into the LLM-compatible textual embedding space:
\begin{equation}
\mathbf{X}_v = \mathcal{P}(\mathbf{Z}_v) \in \mathbb{R}^{N \times D}.
    \label{optim1}
\end{equation}
During our instruction-tuning phase on UAVIT-1M, the language instruction $T$ is embedded as $\mathbf{X}_l \in \mathbb{R}^{M \times D}$. The multimodal fusion occurs within the Transformer blocks of the LLM via the self-attention mechanism, where the concatenated multimodal sequence $\mathbf{X} = [\mathbf{X}_v; \mathbf{X}_l]$ is processed. The cross-modal interaction is formulated as:
\begin{equation}
\text{Attention}(\mathbf{Q}, \mathbf{K}, \mathbf{V}) = \text{softmax}\left(\frac{\mathbf{Q} \mathbf{K}^\top}{\sqrt{d_k}}\right) \mathbf{V},
 \label{optim2}
\end{equation}
where $\mathbf{Q}, \mathbf{K}, \mathbf{V}$ are linear projections of the joint sequence $\mathbf{X}$.
To achieve UAV-specific modality alignment and ensure the model effectively follows complex instructions, we employ an end-to-end auto-regressive training objective. Specifically, given the concatenated multimodal input sequence $\mathbf{X} = [\mathbf{X}_v; \mathbf{X}_l]$, the MLLM is trained to predict the next token in the response sequence $\mathbf{Y} = \{y_1, y_2, \dots, y_L\}$. The fine-tuning process optimizes the negative log-likelihood function conditioned on the multimodal context:
\begin{equation}
\mathcal{L}(\Theta) = - \sum_{i=1}^{L} \log P(y_i \mid \mathbf{X}, \mathbf{Y}_{<i}; \Theta),
 \label{optim3}
\end{equation}
where $L$ is the length of the response, $\mathbf{Y}_{<i}$ denote the preceding tokens in the response, and $\Theta$ represents the trainable parameters.
By combining the alignment layer $\mathcal{P}$ with this loss $\mathcal{L}(\Theta)$, the model achieves a seamless integration of visual perception and language reasoning in the low-altitude aerial scenes.
}

\begin{table*}[t]
\centering
\caption{
Evaluation results of different MLLMs on image-level Cls and VQA tasks.
$\dagger$ denotes results on out-of-domain data. This setting is also termed zero-shot evaluation in many articles, as samples from this domain are unseen during training.
"Pre." represents Presence, "Com." represents Comparison, "Wea." represents Weather condition recognition, "Rec." represents Recognition, "Est" represents Estimation, "int." represents integer, and "dec." represents decimal. 
}
\label{tab:imageclsVQA}
\begin{tabular}{lcccccccccc}
\toprule
\multirow{4}{*}{\textbf{Model}} & \multirow{4}{*}{\textbf{Size}} & \multicolumn{2}{c}{\textbf{Cls} (Acc(\%))}  & \multicolumn{7}{c}{\textbf{VQA} (Acc(\%))} \\
\cmidrule(lr){3-4}  \cmidrule(lr){5-11} 
 &  & \multirow{2}{*}{\textbf{ERA}} & \multirow{2}{*}{\textbf{MOD}$\dagger$} & \multirow{2}{*}{\textbf{Pre.}} &\multirow{2}{*}{\textbf{Com.}}  &\multirow{2}{*}{\textbf{Wea.}} & \textbf{Wea.}$\dagger$ & \multicolumn{3}{c}{\textbf{Flying Altitude}}  \\  \cmidrule(lr){8-8}  \cmidrule(lr){9-11} 
&  &  &  &  &  &  & \footnotesize{UAVDT-S}& Rec. & $\text{Est}_{int.}$ & $\text{Est}_{dec.}$ \\ 
\midrule
\multicolumn{11}{l}{\textit{\textbf{Open-source MLLMs} (unfine-tuned)}} \\
\midrule 
MiniGPT-v2 \cite{chen2023minigpt}  & 7B  &60.34  & 83.13 &74.82 &61.04& 78.62 &89.48 &16.91 &2.05&0.00 \\
LLaVA-1.5 \cite{liu2024improved}    & 7B & 64.37  &82.84 & 81.22 & \textbf{72.21} & \textbf{88.98}&  95.24  & 19.86 &0.00  & 0.00 \\
Qwen-VL-Chat \cite{bai2023qwen}   & 7B   & 73.46  &85.38 & 76.89 & 55.57 &88.03 & 91.95 &35.81 & \textbf{3.93}& \textbf{0.32}\\
InternLM-XC2 \cite{dong2024internlm}  & 7B& 72.81& 81.11 &73.15 & 54.84  & 54.07  &88.57 & 14.21 & 0.27& 0.00  \\
InternVL2.5 \cite{chen2024internvl}   & 8B  & 73.38& 85.51 &78.83  & 69.94 & 55.64 & 91.96 & 36.22& 0.16& 0.00  \\
Qwen2.5-VL \cite{bai2025qwen2}   & 7B  &75.53  &\textbf{96.17} &82.36 & 67.25 & 88.95 &  \textbf{95.83} & \textbf{40.93} &0.21& 0.09  \\
MiniCPM-V-2.6 \cite{yao2024minicpm}   & 8B & \textbf{78.80} & 84.67 & \textbf{83.50} & 66.70 &74.24&  94.91 & 36.48 & 2.44& 0.21\\
DeepSeek-VL2 \cite{wu2024deepseek} & 4.5B  & 77.02  & 93.69 &82.00 &{70.80} & 86.57  & 95.74 &17.21 &2.13& 0.11   \\
GeoChat \cite{kuckreja2024geochat}  & 7B & 64.14 &  83.71 &61.37 & 61.87 &66.74 & 77.98 &36.51 &0.00& 0.00  \\
SkyEyeGPT \cite{zhan2025skyeyegpt}  & 7B &  55.54 & 75.36 &74.50 & 70.16 &67.72& 77.02 &17.05 &0.62&0.00 \\
\midrule
\multicolumn{11}{l}{\textit{\textbf{Closed-source MLLMs} (unfine-tuned)}} \\
\midrule
Gemini 2.5 Flash & $/$  & 84.91 &  97.50 & 87.34 & 77.91& 93.40 &98.88 & 56.45&9.56 &1.06 \\
\midrule
\multicolumn{11}{l}{\textit{\textbf{Low-altitude MLLMs} (fine-tuned on UAVIT-1M)}} \\
\midrule
GeoChat-UAV  & 7B & \underline{88.53}& \textbf{93.49} & {\textbf{90.63}}   & \underline{88.66} & 95.80 & \textbf{99.14} & {\textbf{70.05}}  & \underline{48.10}& \underline{12.43}    \\
$\Delta$  &   & \hgreen{$\uparrow$24.39} & \hgreen{$\uparrow$9.78}& \hgreen{$\uparrow$29.26}   & \hgreen{$\uparrow$26.79} & \hgreen{$\uparrow$29.06}  &  \hgreen{$\uparrow$21.16} & \hgreen{$\uparrow$33.54}  & \hgreen{$\uparrow$48.10}& \hgreen{$\uparrow$12.43}  \\ \cmidrule(lr){1-1} \cmidrule(lr){2-2} \cmidrule(lr){3-11} 
LLaVA1.5-UAV & 7B & 76.52 &83.00  & 86.92 &79.92 & {\textbf{97.00}}& \underline{98.94}& 63.89 & 37.02 & 2.80  \\
$\Delta$  &  &  \hgreen{$\uparrow$12.15} & \hgreen{$\uparrow$0.16} & \hgreen{$\uparrow$5.70} &\hgreen{$\uparrow$7.71} & \hgreen{$\uparrow$8.02}& \hgreen{$\uparrow$3.70} & \hgreen{$\uparrow$44.03} & \hgreen{$\uparrow$37.02} & \hgreen{$\uparrow$2.80} \\ \cmidrule(lr){1-1} \cmidrule(lr){2-2} \cmidrule(lr){3-11} 
MiniGPTv2-UAV  & 7B  &{\textbf{93.06}}  & \underline{85.69}& \underline{89.96} & {\textbf{90.16}} &\underline{96.28} & 97.35 &\underline{66.79} & {\textbf{51.97}} &{\textbf{14.49}} \\
$\Delta$   &    &  \hgreen{$\uparrow$32.72} &\hgreen{$\uparrow$2.56} & \hgreen{$\uparrow$15.14} & \hgreen{$\uparrow$29.12} &\hgreen{$\uparrow$17.66}  &  \hgreen{$\uparrow$7.87} &\hgreen{$\uparrow$49.88} & \hgreen{$\uparrow$49.92} &\hgreen{$\uparrow$14.49} \\ 
\bottomrule
\end{tabular}
\end{table*}

\begin{table*}[!ht]
\renewcommand\tabcolsep{3pt} 
\centering
\caption{
Evaluation results of different MLLMs on image-level D.Cls, and Count tasks. $\dagger$ denotes results on out-of-domain data. 
This setting is also termed zero-shot evaluation in many articles, as samples from this domain are unseen during training.
}
\label{tab:imagedclscount}
\resizebox{1\linewidth}{!}{
\begin{tabular}{lcccccccccccccc}
\toprule
\multirow{4}{*}{\textbf{Model}} & \multirow{4}{*}{\textbf{Size}}  & \multicolumn{6}{c}{\textbf{D.Cls}} & \multicolumn{7}{c}{\textbf{Count} (Acc(\%))}  \\ \cmidrule(lr){3-8} \cmidrule(lr){9-15}
&  & \multicolumn{2}{c}{\textbf{AeroScapes}} & \multicolumn{2}{c}{\textbf{VDD}}& \multicolumn{2}{c}{\textbf{UDD}} & \multirow{2}{*}{\footnotesize{\textbf{AU-AIR}}} & \multirow{2}{*}{\footnotesize{\textbf{\makecell[c]{Drone\\Vehicle}}}} & \multirow{2}{*}{\footnotesize{\textbf{VisDrone}}} & \multicolumn{2}{c}{\small{\textbf{HazyDet}}} &\multirow{2}{*}{\footnotesize{\textbf{UAVDT-M}}} & \multirow{2}{*}{\footnotesize{\textbf{RDDTS}}}$\dagger$ \\ \cmidrule(lr){3-4}  \cmidrule(lr){5-6} \cmidrule(lr){7-8}  \cmidrule(lr){12-13} 
& & \footnotesize{F1(\%)} & \footnotesize{mAP(\%)}& \footnotesize{F1(\%)} & \footnotesize{mAP(\%)} &\footnotesize{F1(\%)} & \footnotesize{mAP(\%)} & & & & \footnotesize{normal} & \footnotesize{hazy} & & \\ 
\midrule
\multicolumn{15}{l}{\textit{\textbf{Open-source MLLMs} (unfine-tuned)}} \\
\midrule
MiniGPT-v2 \cite{chen2023minigpt} & 7B & 10.18&\textbf{9.29}&1.79&27.55&0.98&34.69 & 44.52 & 22.43& 13.20& 16.01 & 15.77  &18.75&20.64 \\
LLaVA-1.5 \cite{liu2024improved}   & 7B  &20.02&7.87&18.18& \textbf{44.36}&13.57 & 27.38 &45.36  & 22.24 & 14.44 & 16.48 &15.53  &25.90& 17.74\\
Qwen-VL-Chat \cite{bai2023qwen}  & 7B   &14.82& 1.18 &22.39&9.07&24.12&14.85 & 50.88 & 27.35&17.33 & 21.14  &20.30  &26.44& 22.66\\
InternLM-XC2 \cite{dong2024internlm}  & 7B &13.67& 0.95&13.34&2.59 &10.45& 3.51  & 44.49 & 23.17& 12.16 & 15.75  & 15.21 &16.07& 14.80\\
InternVL2.5 \cite{chen2024internvl}   & 8B &20.44& 0.82&16.48 & 3.60 &11.16& 4.26 & 46.27 &25.20&17.85 & 21.67  & 20.42 & 22.56& 22.57\\
Qwen2.5-VL \cite{bai2025qwen2}   & 7B  &18.44& 2.06 &23.40& 11.68 &\textbf{26.78}& 17.72 & 50.47 &28.24&18.52 & 21.89  & 21.07 & 25.63& 23.14 \\
MiniCPM-V-2.6 \cite{yao2024minicpm}   & 8B&\textbf{23.46} &2.11 &21.50& 6.84 &24.32& 9.74 & \textbf{51.66} & 24.68& 17.97 & 21.17 & 20.73  &17.61& 21.79\\
DeepSeek-VL2 \cite{wu2024deepseek} & 4.5B &23.35& 3.23 &\textbf{25.36}& 13.79 &20.98& 19.74 & 50.56 & \textbf{28.65}& \textbf{20.64} & \textbf{23.09}  & \textbf{23.04} &\textbf{27.03}& \textbf{23.72} \\
GeoChat \cite{kuckreja2024geochat}   & 7B   &11.50& 1.84 &18.34& 8.99 &18.75& 9.59 &28.05 &16.07&13.37 & 16.69 & 16.97 &8.55&15.43\\
SkyEyeGPT \cite{zhan2025skyeyegpt}  & 7B  &3.17& 4.85 &4.42& 29.36 &1.72& \textbf{44.16} & 33.51 &20.16& 12.99& 15.30  &14.97  &9.42& 18.13\\
\midrule
\multicolumn{15}{l}{\textit{\textbf{Closed-source MLLMs} (unfine-tuned)}} \\
\midrule
Gemini 2.5 Flash & /  & 31.12& 9.83 & 27.44& 50.87 &30.16 & 38.35 & 54.71 &29.94 &25.16 & 26.15 & 26.09 &28.70 & 28.35\\
\midrule
\multicolumn{15}{l}{\textit{\textbf{Low-altitude MLLMs} (fine-tuned on UAVIT-1M)}} \\
\midrule
GeoChat-UAV & 7B &68.62 & \underline{29.15} & \underline{90.08} & \underline{57.18} & \underline{96.53}& \underline{49.05}  & {\textbf{57.33}} & 29.35& {\textbf{26.24}}  & \underline{26.38}  &{\textbf{26.22}}  &25.86& 24.20 \\
$\Delta$ &   & \hgreen{$\uparrow$57.12} & \hgreen{$\uparrow$27.31} & \hgreen{$\uparrow$71.74}& \hgreen{$\uparrow$48.19} & \hgreen{$\uparrow$77.78}& \hgreen{$\uparrow$39.46} & \hgreen{$\uparrow$29.28} & \hgreen{$\uparrow$13.28}& \hgreen{$\uparrow$12.87}  & \hgreen{$\uparrow$9.69}  & \hgreen{$\uparrow$9.25} & \hgreen{$\uparrow$17.31}&\hgreen{$\uparrow$8.77} \\  \cmidrule(lr){1-1} \cmidrule(lr){2-2} \cmidrule(lr){3-15}
LLaVA1.5-UAV & 7B &\underline{72.80}&28.01 &50.67 & 49.03 & 52.00& 44.99  & \underline{57.03} & \underline{31.59} & \underline{25.86} & 23.80 & 23.49  &\underline{27.45}&\underline{24.59}\\
$\Delta$ &   & \hgreen{$\uparrow$52.78} &\hgreen{$\uparrow$20.14} & \hgreen{$\uparrow$32.49} & \hgreen{$\uparrow$4.67} & \hgreen{$\uparrow$38.43} & \hgreen{$\uparrow$17.61}  &\hgreen{$\uparrow$11.67} & \hgreen{$\uparrow$9.35} & \hgreen{$\uparrow$11.42} & \hgreen{$\uparrow$7.32}  & \hgreen{$\uparrow$7.96} &  \hgreen{$\uparrow$1.55}& \hgreen{$\uparrow$6.85}\\ \cmidrule(lr){1-1} \cmidrule(lr){2-2} \cmidrule(lr){3-15}
MiniGPTv2-UAV & 7B   &\textbf{87.73}& {\textbf{39.98}}&\textbf{94.31} & {\textbf{91.33}} &\textbf{98.55} & {\textbf{97.14}} &56.45&{\textbf{32.32}} &25.52 & {\textbf{26.43}} & \underline{26.12} &\textbf{28.03}& {\textbf{25.46}}\\
$\Delta$  &    & \hgreen{$\uparrow$77.55} & \hgreen{$\uparrow$30.69}& \hgreen{$\uparrow$92.52} & \hgreen{$\uparrow$63.78}  & \hgreen{$\uparrow$97.57} &\hgreen{$\uparrow$62.45} &\hgreen{$\uparrow$11.93}&\hgreen{$\uparrow$9.89} &\hgreen{$\uparrow$12.32}& \hgreen{$\uparrow$10.42} & \hgreen{$\uparrow$10.35}  &\hgreen{$\uparrow$9.28} & \hgreen{$\uparrow$4.82}\\
\bottomrule
\end{tabular}
}
\end{table*}

\begin{table*}[t]
\centering
\caption{
 Evaluation results of different MLLMs on the image-level Cap task. "B" represents BLUE, "M." represents METEOR, "R." represents ROUGE, "C." represents CIDER, and "S." represents SPICE.
}
\label{tab:imagecap}
\renewcommand\tabcolsep{3pt} 
\begin{tabular}{lccccccccccccccccc}
\toprule
\multirow{2}{*}{\textbf{Model}} & \multirow{2}{*}{\textbf{Size}}  & \multicolumn{8}{c}{\textbf{capERA}}     & \multicolumn{8}{c}{\textbf{UDV DIT}}    \\
\cmidrule(lr){3-10} \cmidrule(lr){11-18}
 &  & B-1 & B-2 &B-3  & B-4 & M. &R.& C. &S. & B-1 &  B-2 &  B-3 & B-4 & M. &R.& C. &S.\\ 
\midrule
\multicolumn{18}{l}{\textit{\textbf{Open-source MLLMs} (unfine-tuned)}} \\
\midrule
MiniGPT-v2 \cite{chen2023minigpt} & 7B &28.7& 15.0&8.0&4.3&14.7& 25.8&15.4 &11.5& 24.8& 10.7& 4.9 & 2.3&14.0 &22.1& 6.2 &9.2\\
LLaVA-1.5 \cite{liu2024improved}  & 7B & 21.4 &12.6& 7.6&4.6 &16.8& 26.8& 20.3 &13.9 & 20.1& 10.1 & 4.9 &2.3& \textbf{14.8} & 22.7& 11.0 & \textbf{12.7} \\
Qwen-VL-Chat \cite{bai2023qwen} & 7B & 31.9& 17.1  &9.6&5.6& 12.1 & 26.8 & 28.7 & 10.7& 22.8 & 10.6 & 4.9 & 2.3& 11.2& 22.7& 12.0 &7.7\\
InternLM-XC2 \cite{dong2024internlm}  & 7B&18.3 &8.4&5.7& 1.9& 11.8&23.6&13.1 & 4.2& 14.5 & 7.2& 2.8 & 0.6& 10.6  & 16.3 & 2.9 & 2.6\\
InternVL2.5 \cite{chen2024internvl}   & 8B  & 19.7&9.3 &4.4 & 2.1& 14.0& 22.9& 11.5 & 5.3& 15.8 &6.5&2.8 &1.3& 12.3 &17.4&3.4&2.6 \\
Qwen2.5-VL \cite{bai2025qwen2} & 7B &\textbf{36.1} &17.8 &9.1& 4.9& 13.4&28.2& \textbf{30.6} & 11.4& 24.1 & 10.5& 5.4 & 2.8& 12.1 &22.1& 11.9  & 9.5 \\
MiniCPM-V-2.6 \cite{yao2024minicpm}   & 8B & 35.0& 18.8 & 10.7 & 6.2& 14.5&\textbf{28.8}&29.7 & 12.7& \textbf{29.3}& \textbf{12.7}& \textbf{5.9} & \textbf{3.0}& 13.6 &\textbf{23.9} & \textbf{13.1} &11.2 \\
DeepSeek-VL2 \cite{wu2024deepseek} & 4.5B  & 32.2 & 16.6 & 9.2 &5.1&15.1& 28.2& 28.1 &12.3& 23.5 &9.4& 4.4 & 2.0&13.1  &20.5 &10.0&10.7\\
GeoChat \cite{kuckreja2024geochat} & 7B   & 13.7& 6.1& 3.2 &1.9  & 8.2 & 18.7 &12.8 & 4.1& 12.1 &5.5& 2.0 &0.7&8.1 & 18.7&3.7 & 1.5 \\
SkyEyeGPT \cite{zhan2025skyeyegpt}  & 7B  & 30.8& \textbf{18.9}& \textbf{12.3} &\textbf{7.9}& \textbf{16.9} & 28.6&27.4 & \textbf{15.3} &26.7&12.3 &5.5 &2.4& 14.4 &23.8 & 5.0 & 10.7 \\ 
\midrule
\multicolumn{18}{l}{\textit{\textbf{Closed-source MLLMs} (unfine-tuned)}} \\
\midrule
Gemini 2.5 Flash  & /  & 38.8 & 20.1& 14.3& 10.2& 16.8 & 32.3 & 32.9&15.9  & 34.0& 14.5&8.2 & 5.6& 15.7 & 27.7& 15.3 & 14.4 \\ 
\midrule
\multicolumn{18}{l}{\textit{\textbf{Low-altitude MLLMs} (fine-tuned on UAVIT-1M)}} \\
\midrule
GeoChat-UAV  & 7B&\textbf{59.3} &\textbf{46.5}&\textbf{37.9} & \textbf{31.1}& \textbf{25.5}&\textbf{52.2}&\textbf{115.1} &\textbf{24.7}& 51.2 & 33.0& 22.8. &16.1& 18.8  & 39.8 &35.3  &16.6\\
LLaVA1.5-UAV & 7B & 43.0 & 28.4 &20.1&14.7& 15.9& 37.5& 54.9 &15.7& \textbf{65.9}& \textbf{49.9}& \textbf{36.3} & \textbf{25.4}& \textbf{23.2}  & \textbf{53.5} & \textbf{54.2} & \textbf{18.2}\\
MiniGPTv2-UAV & 7B   & 32.2 & 18.9 & 11.6&7.2&16.7 &27.5 &21.6& 14.8& 28.6 &14.7 & 7.4&3.9& 15.8& 26.1& 7.9 & 10.9 \\
\bottomrule
\end{tabular}
\end{table*}

\begin{figure*}[t]
  \centering
  \includegraphics[width=0.9\linewidth]{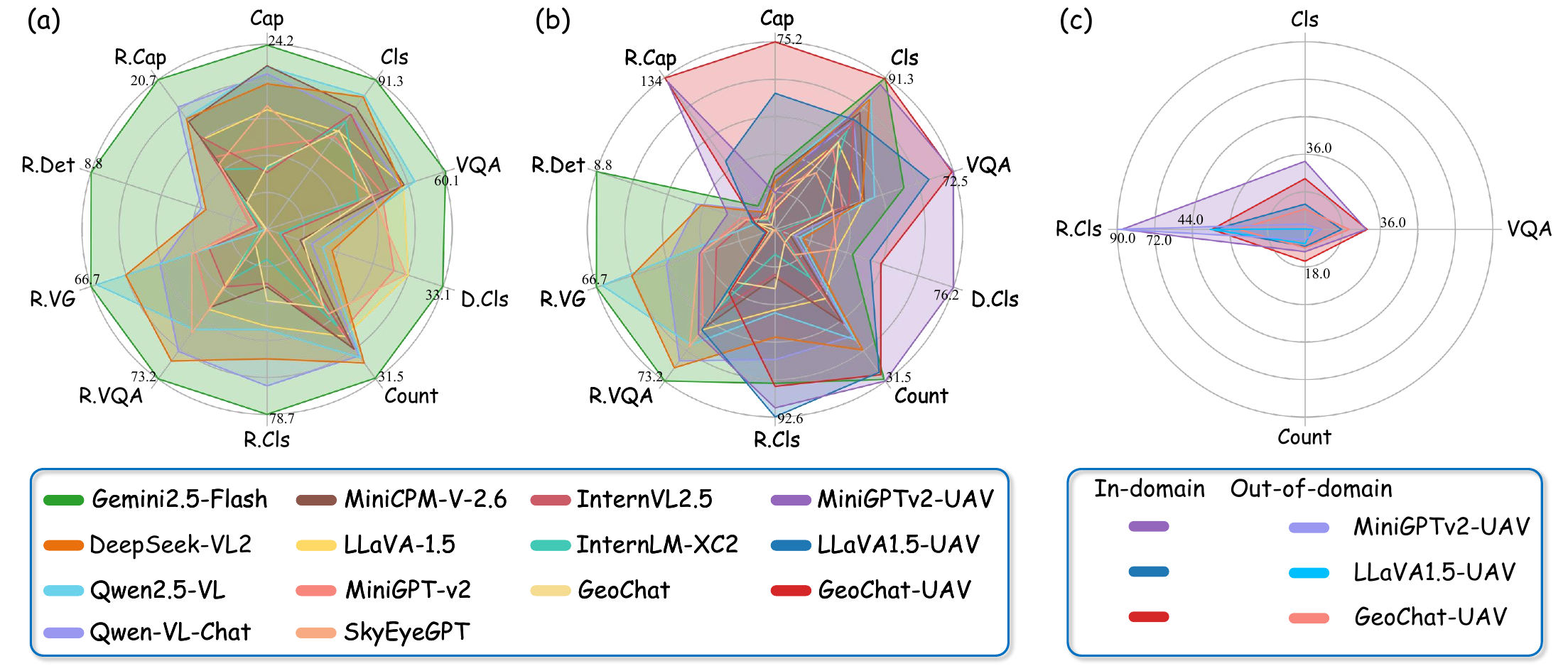}
  \caption{
  (a) Radar charts illustrating performance comparison between 11 open-source MLLMs and closed-source MLLMs across 10 tasks. (b) A comprehensive comparison including the low-altitude MLLMs fine-tuned on our UAVIT-1M.
   (c) Performance gain analysis of the fine-tuned low-altitude MLLMs in both in-domain and out-of-domain tests.
  }
  \label{fig:leida_result}
\end{figure*}

\section{Experiments}

\subsection{Implementation Details}

\subsubsection{Baselines}
We evaluate previous representative MLLMs on our UAVBench benchmark and present an in-depth analysis. 
{We present the performance of the closed-source model without fine-tuning: Gemini 2.5 Flash \cite{comanici2025gemini}.}
We report the performance of SoTA generic open-source MLLMs, including MiniGPTv2 \cite{chen2023minigpt}, LLaVA-1.5 \cite{liu2024improved}, Qwen-VL-Chat \cite{bai2023qwen}, Qwen2.5-VL \cite{bai2025qwen2}, InternLM-XC2 \cite{dong2024internlm}, MiniCPM-V-2.6 \cite{yao2024minicpm}, InternVL2.5 \cite{chen2024internvl}, DeepSeek-VL2 \cite{wu2024deepseek}, and SoTA remote sensing MLLMs, including GeoChat \cite{kuckreja2024geochat} and SkyEyeGPT \cite{zhan2025skyeyegpt}.

\subsubsection{Fine-tuning Hyper-parameters}
To demonstrate the significant value and high quality of our instruction tuning dataset, we provide the comparison results of the MLLMs after fine-tuning on UAVIT-1M. 
We chose LLaVA-1.5 as a model for fine-tuning because it represents a foundational and widely-adopted architecture in the MLLM community.  Numerous academic studies use LLaVA as a starting point for new research. 
By providing a fine-tuned LLaVA1.5-UAV model, we offer an accessible, expandable, and easily comparable baseline that we hope will lay the groundwork for subsequent research in the low-altitude UAV domain.  Similarly, MiniGPTv2 and GeoChat were included to demonstrate our dataset's effectiveness on both general-purpose and domain-specific (remote sensing) architectures.
{All of our experiments are conducted on 4 NVIDIA 48G L20 GPUs. The two L20 GPUs or four NVIDIA RTX 24G 3090 GPUs can also support our experiment.
We detail the structure of the above models and the details of setting the model weights in Table \ref{tab:modelset}.
}

\textbf{MiniGPTv2-UAV:} 
The training details fully follow the open-source code: https://github.com/Vision-CAIR/MiniGPT-4. The image size during fine-tuning is 224$\times$224, the training epoch is 100, and the iterations per epoch are 2000. The fine-tuning stage takes around 35 hours. 
{The learning rate and optimizer settings are set using the default settings from the referenced codebase.}

\textbf{LLaVA1.5-UAV:}
The training parameters fully follow the open-source code: https://github.com/haotian-liu/LLaVA. The image size during fine-tuning is 336$\times$336. The batch size is 8, and the gradient accumulation step is 4. The fine-tuning stage takes around 60 hours.
{The learning rate and optimizer settings are set using the default settings from the referenced codebase.}

\textbf{GeoChat-UAV:}
The training parameters fully follow the open-source code: https://github.com/mbzuai-oryx/GeoChat. 
The image size during fine-tuning is 504$\times$504. The batch size is 8, and the gradient accumulation step is 4.
The fine-tuning stage takes around 70 hours.
{The learning rate and optimizer settings are set using the default settings from the referenced codebase.}

\begin{table*}[!ht]
\centering
\caption{
Evaluation results of different MLLMs on region-level R.VQA and R.Cls tasks. $\dagger$ denotes results on out-of-domain data. 
This setting is also termed zero-shot evaluation in many articles, as samples from this domain are unseen during training.
}
\label{tab:regionvqacls}
\begin{tabular}{lccccccccccc}
\toprule
\multirow{7}{*}{\textbf{Model}} & \multirow{7}{*}{\textbf{Size}} & \textbf{R.VQA}  &\multicolumn{9}{c}{\textbf{R.Cls} (Acc(\%))}  \\ \cmidrule(lr){3-3}  \cmidrule(lr){4-12} 
 &  & \rotatebox{65}{\textbf{UAVDT-S}}$\dagger$ & \rotatebox{65}{\multirow{2}{*}{\textbf{AU-AIR}}} & \rotatebox{65}{\multirow{2}{*}{\textbf{WebUAV-3M}}} & \rotatebox{65}{\multirow{2}{*}{\textbf{DroneVehicle}}} & \rotatebox{65}{\multirow{2}{*}{\textbf{UAVDT-M}}} & \multicolumn{2}{c}{\textbf{HazyDet}} & \rotatebox{65}{\multirow{2}{*}{\textbf{UAVDT-S}}}$\dagger$ & \rotatebox{65}{\multirow{2}{*}{\textbf{RefDrone}}}$\dagger$ &\rotatebox{65}{\multirow{2}{*}{\textbf{RDDTS}}}$\dagger$ \\ 
  &  & Acc(\%) &  & &  & & normal & hazy & & &  \\ 
\midrule
\multicolumn{12}{l}{\textit{\textbf{Open-source MLLMs} (unfine-tuned)}} \\
\midrule
MiniGPT-v2 \cite{chen2023minigpt}  & 7B  & 47.88 &0.00 & 0.00& 0.13 &0.04 & 0.00& 0.00& 0.02 &1.56 & 0.01 \\
LLaVA-1.5 \cite{liu2024improved} & 7B &48.41 & 31.44 & {15.64} & \textbf{76.78} & 55.80& 51.96& 50.43&52.65 & 35.87 & 64.08\\
Qwen-VL-Chat \cite{bai2023qwen}  & 7B  &63.34 & \textbf{75.16} & \textbf{18.43} & 72.12 &\textbf{80.39} &\textbf{80.60} & \textbf{80.80}& \textbf{82.56} & 43.82 & \textbf{85.93} \\
InternLM-XC2 \cite{dong2024internlm} & 7B &37.92 & 20.41 & 6.26 & 39.24 &18.55 &27.15 &27.22 &19.50 & 33.36   & 35.98\\
InternVL2.5 \cite{chen2024internvl}   & 8B  & 40.43& 24.03 & 11.81 & 49.61 &28.16 &32.68 &32.77 & 27.20 & 40.54 & 55.78\\
Qwen2.5-VL \cite{bai2025qwen2}   & 7B &55.57& 58.13 & 12.57 & 69.53 &52.54 & 54.89& 53.71&  49.89  & 37.61 &56.61\\
MiniCPM-V-2.6 \cite{yao2024minicpm}   & 8B &47.54& 54.14 & 7.55 & 45.73 &34.49 & 30.07& 29.89& 30.97 & 40.53   & 39.91\\
DeepSeek-VL2 \cite{wu2024deepseek} & 4.5B & \textbf{66.74}&45.75  & 11.36 & 65.07 & 72.48&75.49 &75.06 &73.67  &38.34  & 78.62 \\
GeoChat \cite{kuckreja2024geochat}  & 7B   & 25.59& 12.74& 14.74& 70.51 &43.89 & 33.99&33.62 & 46.89 & \textbf{53.66} &45.38\\
SkyEyeGPT \cite{zhan2025skyeyegpt}  & 7B & 56.71& 0.12 & 0.01& 0.01 & 0.01&0.01 & 0.00& 0.01 & 2.44 & 0.00  \\
\midrule
\multicolumn{12}{l}{\textit{\textbf{Closed-source MLLMs} (unfine-tuned)}} \\
\midrule
Gemini 2.5 Flash  &  / & 73.11& 87.33 & 27.50  & 89.75 &93.41 & 95.22& 95.31& 79.21 & 51.63 &88.64 \\
\midrule
\multicolumn{12}{l}{\textit{\textbf{Low-altitude MLLMs} (fine-tuned on UAVIT-1M)}} \\
\midrule
GeoChat-UAV & 7B &30.56 & 68.76 & \textbf{87.44} & \textbf{97.46} &80.94 &71.22 & 71.58 &85.37& 68.22 & 88.40\\
$\Delta$ &  &\hgreen{$\uparrow$4.97} & \hgreen{$\uparrow$56.02} & \hgreen{$\uparrow$72.70} & \hgreen{$\uparrow$26.95} &\hgreen{$\uparrow$37.05} &\hgreen{$\uparrow$37.23} &\hgreen{$\uparrow$37.96} &  \hgreen{$\uparrow$38.48} & \hgreen{$\uparrow$14.56} & \hgreen{$\uparrow$43.02} \\ \cmidrule(lr){1-1} \cmidrule(lr){2-2} \cmidrule(lr){3-12} 
LLaVA1.5-UAV & 7B  & \underline{48.48}& \underline{84.89} & 74.05 & \underline{91.89} &\textbf{95.36} & \textbf{98.68}&\textbf{98.80}& \textbf{96.72} & {\textbf{93.80}} &{\textbf{98.45}}\\
$\Delta$ &   &\hgreen{$\uparrow$0.07} &\hgreen{$\uparrow$53.45} & \hgreen{$\uparrow$58.41} & \hgreen{$\uparrow$15.11} &\hgreen{$\uparrow$39.56} &\hgreen{$\uparrow$46.72} & \hgreen{$\uparrow$48.37}& \hgreen{$\uparrow$44.07} & \hgreen{$\uparrow$57.93}  &\hgreen{$\uparrow$34.37} \\ \cmidrule(lr){1-1} \cmidrule(lr){2-2} \cmidrule(lr){3-12} 
MiniGPTv2-UAV& 7B   & \textbf{50.65} &{\textbf{84.99}} &\underline{83.83} & 84.07 &\underline{89.13} &\underline{92.35} &\underline{93.20} & \underline{92.91} & \underline{86.87} & \underline{92.09}  \\
$\Delta$ &   & \hgreen{$\uparrow$2.77} &\hgreen{$\uparrow$84.99} &\hgreen{$\uparrow$83.83} & \hgreen{$\uparrow$83.94} & \hgreen{$\uparrow$89.09}& \hgreen{$\uparrow$92.35} &\hgreen{$\uparrow$93.20} & \hgreen{$\uparrow$92.89}& \hgreen{$\uparrow$85.31} & \hgreen{$\uparrow$92.08}  \\
\bottomrule
\end{tabular}
\end{table*}

{
\subsubsection{Data Balance Sampling Strategy}
The weight sampling strategy is used to assign different sampling weights to each task, maintaining data balance by adjusting the sampling frequency.
Calculate the proportion based on the sample size of each task, and then assign an appropriate sampling frequency according to the proportion. The higher the proportion, the smaller the sampling frequency.
This strategy allows our fine-tuned MLLMs to fairly gain the ability to solve each task, ensuring that large-scale data will not dominate the training process and small-scale data will not be ignored.
}

\subsection{Main Results on UAVBench}

\subsubsection{Image Understanding of Low-altitude UAV Scenarios}
We provide complete evaluation results of five image-level tasks: Cls and VQA tasks in Table \ref{tab:imageclsVQA}, D.Cls and Count tasks in Table \ref{tab:imagedclscount}, and Cap task in Table \ref{tab:imagecap}.
Our analysis reveals that among the open-source models, the more recent MLLMs with advanced general capabilities, such as Qwen2.5-VL, DeepSeek-VL2, and MiniCPM-V-2.6, consistently exhibit more prominent performance across most tasks. 
As shown in Figure \ref{fig:leida_result} (a), the top-performing open-source model for each of the five image-level tasks is as follows: Qwen2.5-VL for Cls and VQA, LLaVA-1.5 for D.Cls, DeepSeek-VL2 for Count, and MiniCPM-V-2.6 for Cap.

However, these open-source MLLMs face significant difficulties with more specialized and complex low-altitude tasks.
This is particularly evident in the VQA task for flying altitude, where nearly all open-source models fail to provide meaningful answers for altitude estimation, with most scores being 0.00\%. 
Furthermore, D.Cls and Count tasks that require a more granular understanding of the entire image remain a considerable challenge.
As shown in Table \ref{tab:imagedclscount}, the mAP and F1 scores for D.Cls are generally low for open-source MLLMs, and the accuracy for counting objects in complex scenes is modest at best.
In contrast, the closed-source model consistently outperforms all open-source MLLMs on most image-level tasks. It also struggles with the difficult altitude estimation task. Its scores of 9.56\% (integer) and 1.06\% (decimal) are significantly better than the near-zero scores from the open-source models, highlighting the difficulty of this specialized UAV-centric task.

\begin{table*}[t]
\centering
\caption{
Evaluation results of different MLLMs on region-level Det and VG tasks. $\dagger$ denotes results on out-of-domain data. 
This setting is also termed zero-shot evaluation in many articles, as samples from this domain are unseen during training.
}
\label{tab:regiondetvg}
\begin{tabular}{lccccccccccc}
\toprule
\multirow{7}{*}{\textbf{Model}} & \multirow{7}{*}{\textbf{Size}}  &\multicolumn{8}{c}{\textbf{Det} (AP@IoU=0.5(\%))} &\multicolumn{2}{c}{\textbf{VG} (Acc@0.5(\%))} \\ \cmidrule(lr){3-10} \cmidrule(lr){11-12} 
 &   & \rotatebox{65}{\multirow{2}{*}{\textbf{AU-AIR}}} & \rotatebox{65}{\multirow{2}{*}{\textbf{DroneVehicle}}} &\rotatebox{65}{\multirow{2}{*}{\textbf{UAVDT-M}}} & {\rotatebox{65}{\multirow{2}{*}{\textbf{VisDrone}}}} & \multicolumn{2}{c}{\textbf{HazyDet}} & {\rotatebox{65}{\multirow{2}{*}{\textbf{RefDrone}}}}$\dagger$ & {\rotatebox{65}{\multirow{2}{*}{\textbf{RDDTS}}}}$\dagger$& {\rotatebox{65}{\multirow{2}{*}{\textbf{RefDrone}}}}$\dagger$ &\rotatebox{65}{\multirow{2}{*}{\textbf{UAVDT-S}}}$\dagger$\\ 
  & & & & & & {normal}& {hazy} & & & & \\ 
\midrule
\multicolumn{12}{l}{\textit{\textbf{Open-source MLLMs} (unfine-tuned)}} \\
\midrule
MiniGPT-v2 \cite{chen2023minigpt}  & 7B  &4.21& 0.35 & 0.47 &0.24 & 0.52& 0.48 & 0.02 &0.72 &12.46 &42.18\\
LLaVA-1.5 \cite{liu2024improved} & 7B & 0.46& 0.04 & 0.00 & 0.00& 0.04&0.03 & 0.03& 0.02 & 2.09 & 2.90\\
Qwen-VL-Chat \cite{bai2023qwen}  & 7B  & {13.60}& {3.04} & {3.90}& {0.35}& {1.24}& {1.34} &{1.03}&1.84& 29.53& {51.50}\\
InternLM-XC2 \cite{dong2024internlm} & 7B &0.94& 0.96  & 0.62& 0.10 & 0.04 &0.03& 0.08 & 0.21& 2.49 & 3.34 \\
InternVL2.5 \cite{chen2024internvl}   & 8B  & 1.05 &1.31& 0.71& 0.15 & 0.04 & 0.04 & 0.11& 0.57& 5.66 &38.21 \\
Qwen2.5-VL \cite{bai2025qwen2}   & 7B & \textbf{14.95}&\textbf{5.97}  & \textbf{5.54}& \textbf{1.55} & \textbf{3.65} &\textbf{3.70}&\textbf{2.21}& \textbf{4.65}  & {\textbf{69.08}} & {\textbf{60.08}} \\
DeepSeek-VL2 \cite{wu2024deepseek} & 4.5B  &14.17 & {2.03}& 1.08& 0.89 & {1.56} & {1.53}& 1.81 & {1.44}& {56.10} &50.94 \\
GeoChat \cite{kuckreja2024geochat}  & 7B  & 0.01 & 0.01 & 0.00& 0.00 & 0.00&0.00 & 0.01&0.00& 0.52 &0.56\\
SkyEyeGPT \cite{zhan2025skyeyegpt}  & 7B & 2.28 &0.38  & 0.59 & 0.25 &0.45& 0.58& 0.00 & 0.35& 12.37 & 44.63\\
\midrule
\multicolumn{12}{l}{\textit{\textbf{Closed-source MLLMs} (unfine-tuned)}} \\
\midrule
Gemini 2.5 Flash  &  / &19.26  & 10.42  &  8.98 & 5.07 & 7.33& 7.30 & 5.80 & 6.14& 73.32 & 65.81\\
\midrule
\multicolumn{12}{l}{\textit{\textbf{Low-altitude MLLMs} (fine-tuned on UAVIT-1M)}} \\
\midrule
GeoChat-UAV & 7B & 0.29& 0.23  & 0.01 & 0.10 &0.01 & 0.00 & 0.10 & 0.02 & 1.66 & 4.38 \\
$\Delta$ &  & \hgreen{$\uparrow$0.28}& \hgreen{$\uparrow$0.22}  & \hgreen{$\uparrow$0.01} & \hgreen{$\uparrow$0.10} &\hgreen{$\uparrow$0.01} & \hgreen{0.00} &\hgreen{$\uparrow$0.10}& \hgreen{$\uparrow$0.02} & \hgreen{$\uparrow$1.14} & \hgreen{$\uparrow$3.82} \\ \cmidrule(lr){1-1} \cmidrule(lr){2-2} \cmidrule(lr){3-12} 
LLaVA1.5-UAV & 7B  & 1.29& 0.12 &0.02 &0.01 & 0.33 & 0.26&0.04& 0.00 & 2.09 & 5.37 \\
$\Delta$ &   & \hgreen{$\uparrow$0.83}& \hgreen{$\uparrow$0.08}  &\hgreen{$\uparrow$0.02} &\hgreen{$\uparrow$0.01} & \hgreen{$\uparrow$0.29} & \hgreen{$\uparrow$0.23} &\hgreen{$\uparrow$0.01}& \hblue{$\downarrow$0.02} & \hgreen{0.00} & \hgreen{$\uparrow$2.47} \\ \cmidrule(lr){1-1} \cmidrule(lr){2-2} \cmidrule(lr){3-12} 
MiniGPTv2-UAV& 7B   & \textbf{7.09} & \textbf{1.47}& \textbf{1.27} &\textbf{0.28}& \textbf{0.99} & \textbf{0.91} &\textbf{0.17}& \textbf{0.72} & \textbf{13.76}&\textbf{42.78}\\
$\Delta$ &   & \hgreen{$\uparrow$2.88} & \hgreen{$\uparrow$1.12} & \hgreen{$\uparrow$0.80} &\hgreen{$\uparrow$0.04} & \hgreen{$\uparrow$0.47} & \hgreen{$\uparrow$0.43} &\hgreen{$\uparrow$0.15}& \hgreen{0.00} & \hgreen{$\uparrow$1.30}&\hgreen{$\uparrow$0.60}\\
\bottomrule
\end{tabular}
\end{table*}

\begin{table*}[!ht]
\centering
\caption{
  Evaluation results of different MLLMs on the region-level R.Cap task.
  $\dagger$ denotes results on out-of-domain data. 
This setting is also termed zero-shot evaluation in many articles, as samples from this domain are unseen during training.
  "B" represents BLUE, "M." represents METEOR, "R." represents ROUGE, "C." represents CIDER, and "S." represents SPICE. 
}
\label{tab:regioncap}
\renewcommand\tabcolsep{3pt} 
\begin{tabular}{lccccccccccccccccc}
\toprule
\multirow{2}{*}{\textbf{Model}} & \multirow{2}{*}{\textbf{Size}}  & \multicolumn{8}{c}{\textbf{WebUAV-3M}}     & \multicolumn{8}{c}{\textbf{UAVDT-S}$\dagger$}    \\
\cmidrule(lr){3-10} \cmidrule(lr){11-18}
 &  & B-1 & B-2 &B-3  & B-4 & M. &R.& C. &S. & B-1 &  B-2 &  B-3 & B-4 & M. &R.& C. &S.\\ 
\midrule
\multicolumn{18}{l}{\textit{\textbf{Open-source MLLMs} (unfine-tuned)}} \\
\midrule
MiniGPT-v2 \cite{chen2023minigpt} & 7B &0.1&0.1&0.0&0.0&2.6& 7.0&10.0 &7.0& \textbf{54.2}& \textbf{42.5}& \textbf{43.6}& 0.0&24.6&\textbf{62.8}& \textbf{281.1} &\textbf{59.1}\\
LLaVA-1.5 \cite{liu2024improved}  & 7B & 5.4 & 2.8& 1.3 & 0.6 & 5.0& 13.9& 12.4& 6.7& 33.6  & 21.2 &15.5 & 0.0& 18.0 &42.9&39.5 & 27.0  \\
Qwen-VL-Chat \cite{bai2023qwen} & 7B & 11.2& 5.9 &\textbf{3.1}&\textbf{1.8} &6.6 & 17.8 & \textbf{16.9} &15.3& 39.1& 23.0& 11.7 & 0.0& 24.0&45.6& 29.8 & 40.6 \\
InternLM-XC2 \cite{dong2024internlm}  & 7B& 1.8& 1.0& 0.3 & 0.0 & 2.5 & 5.6 & 8.3 & 4.5 & 20.6& 13.4 & 5.6 & 0.0& 10.7 & 22.0 & 10.8 & 15.7 \\ 
InternVL2.5 \cite{chen2024internvl}   & 8B  & 2.2& 1.1 & 0.5 &0.2 & 4.7 & 10.8 & 12.9 & 9.1 & 25.2& 16.7 & 6.9 & 0.0& 11.4 & 25.6 & 11.3 & 19.3 \\ 
Qwen2.5-VL \cite{bai2025qwen2}   & 7B   & \textbf{12.6}& \textbf{6.3} & 2.5  & 1.6 & \textbf{8.1} & \textbf{18.5} & {15.4} &\textbf{17.2}& 41.6 &26.4& 12.1 & 0.0&\textbf{25.8} & 47.3 & 35.1 & 42.1 \\
MiniCPM-V-2.6 \cite{yao2024minicpm}   & 8B &3.6& 1.9 & 1.0&0.5&4.7 &11.8 &14.9 &9.1& 28.5&19.4 & 8.2 & 0.0& 15.1 & 37.3 & 22.9 & 32.5 \\ 
DeepSeek-VL2 \cite{wu2024deepseek} & 4.5B  & 11.9&5.7 & 2.8 & 1.7  & 5.4  & 16.8 & 15.3 & 13.4& 24.6 & 20.1& 12.3 & 0.0& 22.4 & 35.8 & 25.7 & 38.6 \\
GeoChat \cite{kuckreja2024geochat} & 7B   &2.4&1.0 & 0.5  & 0.1  & 3.4 & 9.1&3.8 &5.9& 0.7 &0.0& 0.0 & 0.0& 2.9 & 0.9& 0.6 &1.1 \\
SkyEyeGPT \cite{zhan2025skyeyegpt} & 7B  & 0.0& 0.0 & 0.0 & 0.0 & 2.4 & 5.9 & 9.1 &6.9 & 46.6& 39.4 & 42.1 & 0.0& 23.1 &53.7 & 257.0 & 56.6 \\ 
\midrule
\multicolumn{18}{l}{\textit{\textbf{Closed-source MLLMs} (unfine-tuned)}} \\
\midrule
Gemini 2.5 Flash& /  & 14.8& 9.6 & 5.9 & 3.0 & 10.1& 21.3 & 20.7 & 20.9 & 40.8& 31.6 & 22.4 & 0.0 & 28.0 & 52.4 & 50.6 & 55.3 \\ 
\midrule
\multicolumn{18}{l}{\textit{\textbf{Low-altitude MLLMs} (fine-tuned on UAVIT-1M)}} \\
\midrule
GeoChat-UAV  & 7B   & \textbf{49.9}& \textbf{35.5} &\textbf{25.4}  & \textbf{18.6}  & \textbf{21.2} & \textbf{45.9} & \textbf{133.6} &\textbf{24.5}& 8.3 & 0.6& 0.0 & 0.0& 12.5 & 13.1 & 1.7 & 3.4 \\
LLaVA1.5-UAV  & 7B & 45.9 & 31.5& 20.9 & 13.9& 17.1& 41.1& 60.1 &12.7 & 8.5  & 2.8 & 0.0 &0.0 & 10.2 & 14.4& 1.6 & 6.7 \\
MiniGPTv2-UAV & 7B   &46.4 & 32.9 & 23.5&17.1&20.1&45.6 &130.5& 24.0 & 8.8 & 2.1& 0.6&0.0&12.4 & 14.7& 1.7 & 6.1 \\
\bottomrule
\end{tabular}
\end{table*}

\begin{table*}[t]
\centering
\caption{
{The scoring criteria of the four-level rating system for the generated region-level captions.}
}
\label{tab:hunman_evaluation}
\begin{tabular}{lll}
\toprule
\textbf{Dimension} & \textbf{Level} & \textbf{Criteria} \\  \midrule
\multirow{4}{*}{Object} &
  A &
  The caption has the correct types of objects and rich attribute information. \\
& B &
  The caption has objects and attribute information, with an accuracy higher than 50\%.  \\
& C &
  The caption has objects and attribute information, but with an accuracy less than 50\%.  \\
& D &
  The caption has no object descriptions. \\  \midrule
\multirow{4}{*}{Position} &
  A &
  The caption has correct position descriptions for objects. \\
& B &
  The caption has position descriptions for objects with an accuracy higher than 50\%. \\
& C &
  The caption has position descriptions, but with an accuracy less than 50\%. \\
& D &
  The caption has no position descriptions. \\ \midrule
\multirow{4}{*}{Hallucination} &
  A &
  The caption has no hallucination description. \\
& B &
  The caption has hallucination description, and it accounts for less than 50\% \\
& C &
  The caption has a large proportion of hallucination description, more than 50\% \\
& D &
  The caption is entirely hallucination description. \\ 
  \bottomrule
\end{tabular}
\end{table*}

\begin{figure*}[t]
  \centering
  \includegraphics[width=0.99\linewidth]{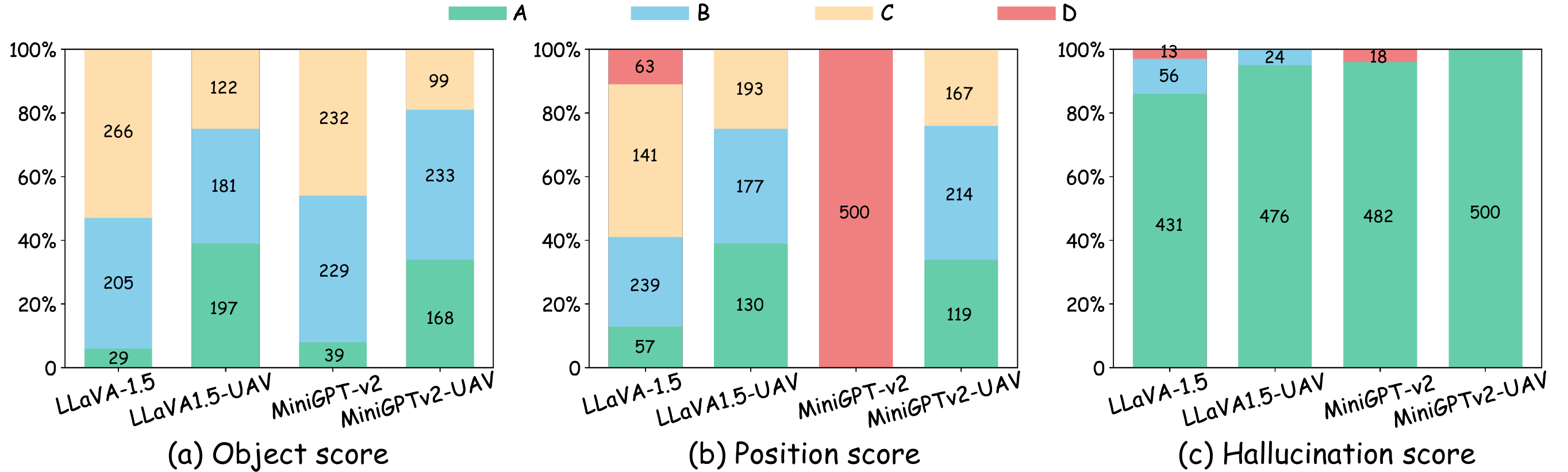}
  \caption{
  The comparison among LLaVA-1.5, LLaVA1.5-UAV, MiniGPT-v2, and MiniGPTv2-UAV on the UAVDT-S test unit of the region caption task. The scores are manually evaluated from three dimensions by 10 volunteers, as shown in (a), (b), and (c).
  }
  \label{fig:human_caption}
\end{figure*}

\subsubsection{Region Perception of Low-altitude UAV Scenarios}
We present the complete evaluation results for the five region-level tasks: R.VQA and R.Cls tasks in Table \ref{tab:regionvqacls}, Det and VG tasks in Table \ref{tab:regiondetvg}, and R.Cap task in Table \ref{tab:regioncap}.
R.VQA and R.Cls tasks require a general understanding of a region in images. As shown in Table \ref{tab:regionvqacls}, leading open-source models like DeepSeek-VL2 and Qwen-VL-Chat achieve competitive accuracy, demonstrating their ability to classify objects and answer questions about specific image regions.
As shown in Figure \ref{fig:leida_result} (a), the closed-source model continues to achieve the highest scores in these tasks.

 \textbf{Region-level Detection and Grounding Task.}
UAVBench is particularly challenging for the low-altitude object detection task. As detailed in Table \ref{tab:regiondetvg}, most open-source models struggle to achieve 5\% AP@0.5 over all 8 test units.
This stands in stark contrast to their performance on benchmarks of the natural domain, suggesting that the skills honed on internet-scale data do not transfer to the low-altitude aerial scenes.
Qwen2.5-VL and DeepSeek-VL2 perform the most prominently in visual grounding, but remain limited and lack practical usability.
This failure in localization represents a significant bottleneck. It suggests that while MLLMs can often understand what is in a region, they lack the fine-grained perception to identify exactly where it is.
This deficiency is a critical barrier to deploying these models in real-world UAV applications such as search and rescue, infrastructure inspection, or autonomous navigation, where precision is paramount.
The difficulty of UAVBench comes from its inherent characteristics, which are out-of-distribution for Internet data.
Unlike the ground-level, object-centric images common in pre-training datasets, our UAVBench features images with drastic scale variations, unconventional bird's-eye perspectives, and a high density of small objects.
These factors make precise localization a far more demanding task. 
Even the closed-source model, despite outperforming its open-source models, demonstrates the universal difficulty of these tasks.

\subsection{Fine-tuning Results using UAVIT-1M}
We showcase the superior quality of UAVIT-1M by presenting the fine-tuning results at the bottom of Tables \ref{tab:imageclsVQA} - \ref{tab:regioncap}. LLaVA1.5-UAV, MiniGPTv2-UAV, and GeoChat-UAV establish new records on many test units.
The results show that adding UAVIT-1M increases the overall performance on UAVBench at all levels of tasks. 
As shown in Figure \ref{fig:leida_result} (b), for image-level tasks, fine-tuned low-altitude models even surpass the advanced closed-source model.
For region-level tasks, low-altitude models can achieve the best performance on R.cls and R.Cap tasks. However, for R.VQA, Det, and VG tasks, the performance gain is still very limited. 
The models show different strengths. The closed-source model excels at Cls, Count, R.VQA, Det, and VG. MiniGPTv2-UAV is particularly adept at VQA, D.Cls, Count, R.Cls, and R.Cap, while GeoChat-UAV performs best in Cap, Cls, VQA, and R.Cap. 
In addition, the models fine-tuned on UAVIT-1M demonstrate superior performance on both in-domain and out-of-domain data.
As illustrated in Figure \ref{fig:leida_result} (c), the fine-tuned models exhibit a more significant performance gain in-domain compared to out-of-domain.
A comparative analysis reveals that the radar charts for the same model maintain a similar shape across both domains, indicating a consistent distribution of performance gains across different tasks.
This suggests that the fine-tuning process not only enhances the models' ability on familiar data but also improves their generalization to unseen scenarios, although it has a more obvious effect on data within the training distribution.

\begin{figure*}[t]
  \centering
  \includegraphics[width=0.99\linewidth]{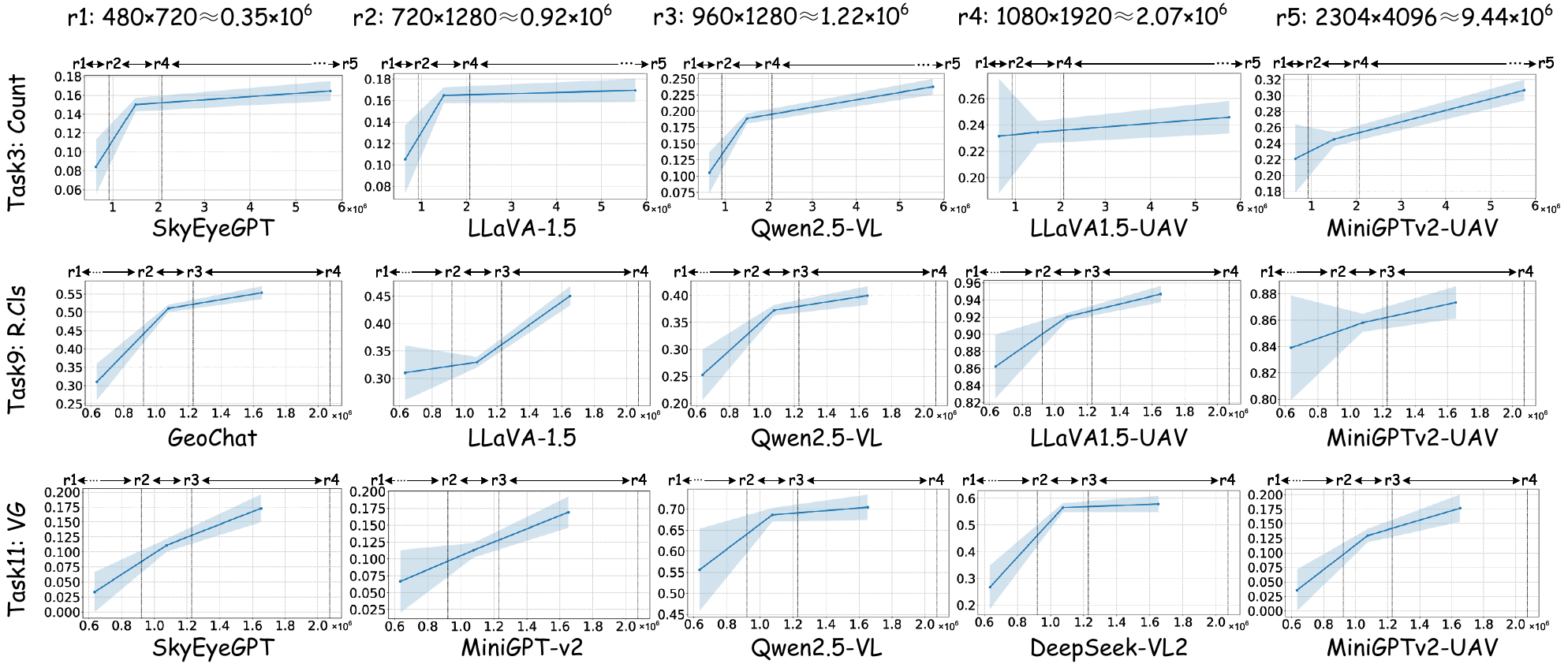}
  \caption{
  {Performance comparisons of some representative or SOTA models at different image resolutions and on the Count, R.Cls, and VG tasks. The resolutions r1 through r5 correspond to mainstream video resolutions, namely 480P, 720P, 960P, 1080P, and 4K, respectively. Obviously, there is a positive correlation between image resolution and model performance.
  }
  }
  \label{fig:resolu}
\end{figure*}

\begin{figure*}[t]
  \centering
  \includegraphics[width=0.89\linewidth]{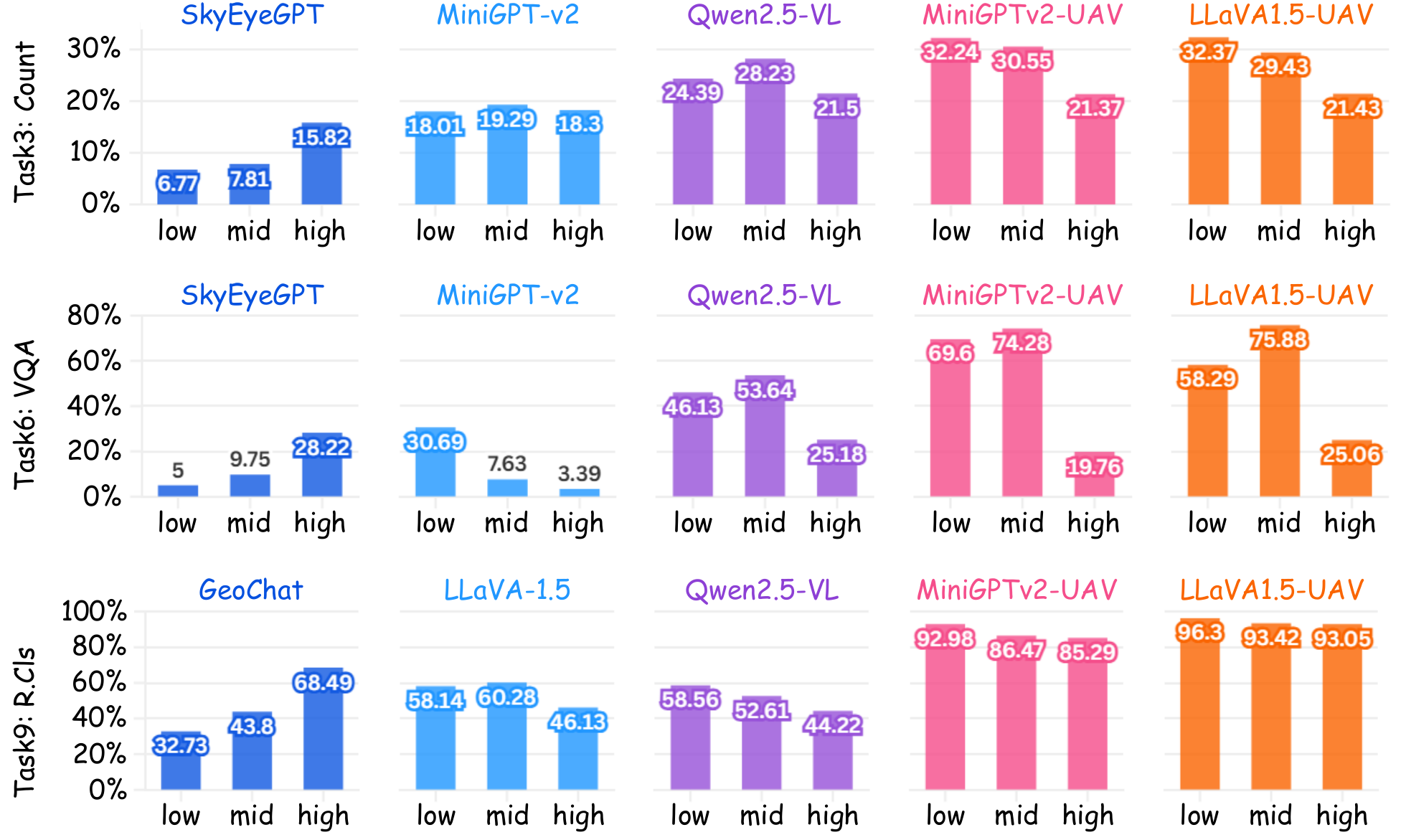}
  \caption{
  {Performance comparisons of some representative or SOTA models at different flight altitudes and on the Count, VQA, and R.Cls tasks.
  Flying altitude indicates the flying height of UAVs when the images are captured, \textit{i.e.}, low ($<$ 30m), mid ($\geq$ 30m and $\leq$ 70m), and high ($>$ 70m).
  }
  }
  \label{fig:flyaltitu}
\end{figure*}

\begin{figure*}[t]
  \centering
  \includegraphics[width=0.89\linewidth]{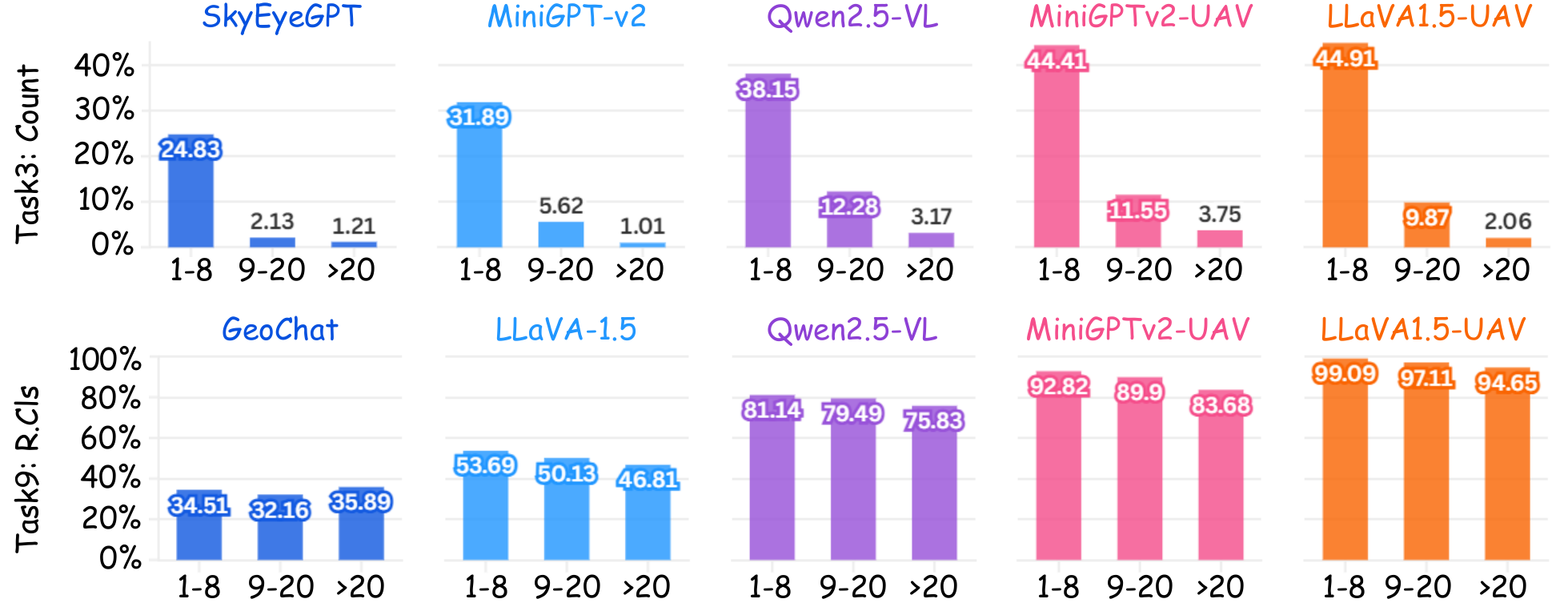}
  \caption{
  {Performance comparisons of some representative or SOTA models at different target quantities and on the Count and R.Cls tasks.
  }
  }
  \label{fig:objectnumber}
\end{figure*}

 \textbf{Human Evaluation for the R.Cap Task.}
A notable difference arises in the Region Captioning task. While the metrics on the UAVDT-S test unit in Table \ref{tab:regioncap} suggest poor performance, this is largely a data issue rather than a methodological issue. 
The ground-truth annotations for this dataset consist of short phrases derived from the corresponding visual grounding task, rather than descriptive sentences. 
The metrics that rely on n-gram overlap, such as BLEU and CIDEr, unfairly evaluate the more fluent, sentence-based captions generated by the MLLMs. Consequently, we conduct human evaluations. 
{
Specifically, we selected 50 representative images and the corresponding object regions from the UAVDT-S test unit.
To ensure these results generalize, the images were carefully sampled to cover a wide distribution of scenarios, including varying weather conditions, flight altitudes, camera views, and object densities.
Previous work \cite{hu2025rsgpt,yao2025falcon} scores the generated image captions from three dimensions: detail description, position description, and hallucination description. Differently, our region-level captions are evaluated across three dimensions: object, position, and hallucination. Each dimension is rated using the four-level scoring system of A, B, C, and D \cite{hu2025rsgpt,yao2025falcon}.
We established a four-level scoring system (A, B, C, D) to maintain consistency across reviewers. The specific criteria for each level are as shown in Table \ref{tab:hunman_evaluation}.
We invited 10 volunteers to assess the generated region captions. To ensure objectivity, the evaluation was conducted in a blind setting: annotators were provided only with the source images, the object regions, and the generated captions, without knowing which model generated which caption.
}

As shown in Figure \ref{fig:human_caption}, among all three dimensions, LLaVA1.5-UAV and MiniGPTv2-UAV received the fewest C and D ratings, while the most A and B ratings.
They not only avoid the hallucination issues but also describe the object and location information more accurately. It is not the performance degradation indicated by the decline of the caption metrics in Table \ref{tab:regioncap}.
In the position score, MiniGPT-v2 only obtains D. This is because the region-level captions of MiniGPT-v2 only contain object nouns and cannot generate position-related descriptions.
The results show that, compared with the original model, the score of the LLaVA1.5-UAV and MiniGPTv2-UAV has significantly improved.

\begin{table*}[t]
\centering
\caption{
{
Quantitative analysis of the domain gap between UAV, Natural, and Satellite RS images. The metrics reported represent the mean values and variance across 20 independent trials.
}
}
\label{tab:RSuavdis}
\begin{tabular}{ccc|ccc}
\toprule
\multicolumn{3}{c}{Fréchet Inception Distance}    & \multicolumn{3}{|c}{Cosine Similarity} \\
\midrule
UAV $\leftrightarrow$ RS  & UAV $\leftrightarrow$ Natural & RS $\leftrightarrow$ Natural & UAV $\leftrightarrow$ RS   & UAV $\leftrightarrow$ Natural   & RS $\leftrightarrow$ Natural  \\
0.60$\pm$0.002 & 0.57$\pm$0.004   & 0.73$\pm$0.007  &  0.73$\pm$0.002  &  0.81$\pm$0.001    &0.72$\pm$0.002    \\
\bottomrule
\end{tabular}
\end{table*}

\begin{figure*}[t]
  \centering
  \includegraphics[width=0.97\linewidth]{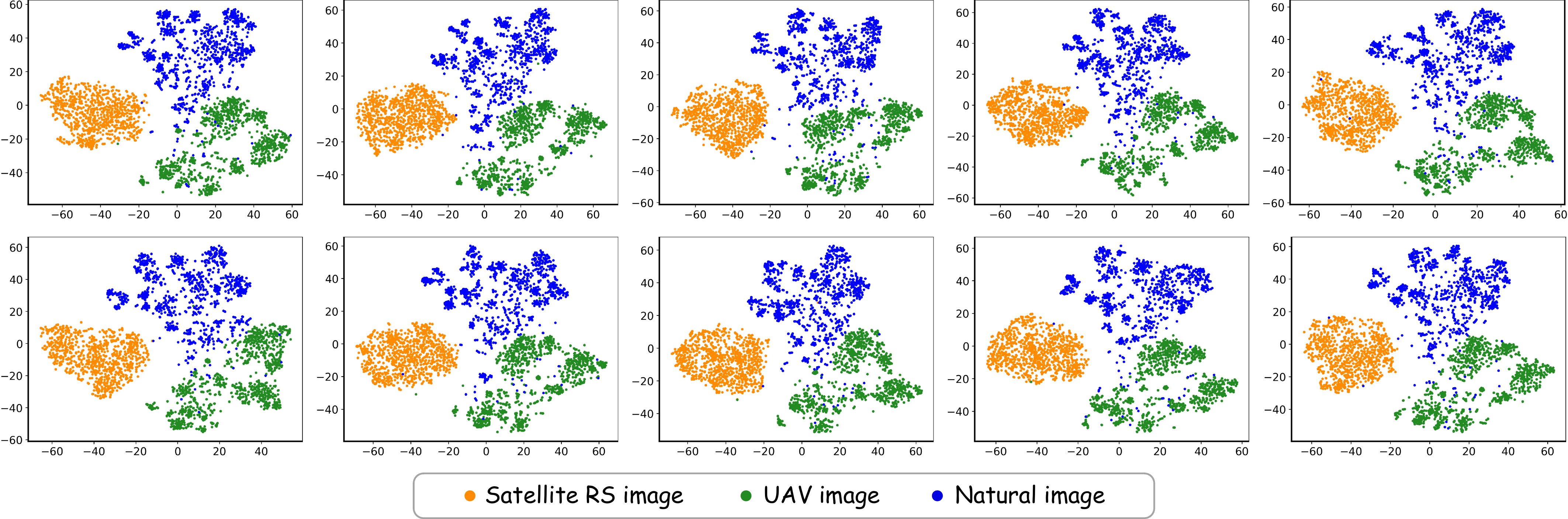}
  \caption{
  {Visualization of the domain gap between low-altitude UAV data and natural/satellite RS data using t-SNE clustering on visual features.
  Here, only the results of 10 sampling are presented.
  }
  }
  \label{fig:T-SNE}
\end{figure*}

\subsection{Detailed Analysis}

{
\textbf{Performance variation across task categories.}
As illustrated in Figure \ref{fig:leida_result} (b), which showcases the performance of various MLLMs across 10 low-altitude tasks, a clear hierarchical trend in task difficulty emerges. Based on the best accuracy achieved for each task, the average performance ranks from highest to lowest as follows: R.Cls $>$ Cls $>$ D.Cls $>$ R.VQA $>$ VQA $>$ R.VG $>$ Count $>$ R.Det. 
This reveals that classification-based tasks reside in the first performance tier, as they primarily require the model to map global or local visual features to predefined semantic categories. While standard image classification (Cls) is relatively straightforward, detailed classification (D.Cls) presents a higher challenge as it necessitates a granular understanding of all entities within a scene. 
VQA-based tasks, including region VQA (R.VQA) and image VQA, constitute the second tier. Their complexity stems from the need to align multi-level visual cues with language queries. 
The most significant performance bottlenecks occur in visual grounding (R.VG), target counting (Count), and region detection (R.Det). These tasks represent the highest level of difficulty because they demand precise spatial coordinate regression and fine-grained numerical reasoning. 
In low-altitude UAV scenarios, factors such as scale variations, small object densities, and dynamic perspectives further increase the difficulty of these tasks.
}

{
\textbf{Performance biases across image resolutions.}
As illustrated in Figure \ref{fig:resolu}, we conduct a statistical analysis of the average performance across different image resolution intervals for SOTA generic MLLMs, RS-MLLMs, and our fine-tuned low-altitude MLLMs. A definitive positive correlation between input image resolution and model accuracy is observed across three representative tasks: Target Counting, Region Classification, and Visual Grounding. Specifically, the resolutions $r1$ through $r5$ correspond to mainstream video standards, namely 480P, 720P, 960P, 1080P, and 4K, respectively.
The results reveal that most models experience their most significant performance surge when moving from $r1$ to $r3$. This sharp upward trend suggests that increasing resolution at lower tiers effectively restores critical semantic details and fine-textured cues. However, the rate of improvement notably plateaus as resolution advances from $r3$ to $r5$.
This diminishing marginal utility likely stems from the fixed patch-size processing inherent in vision encoders. When the pixel density exceeds a certain threshold, further increases will add very little additional information to the visual tokens.
}

{
\textbf{Performance biases across flight altitudes.} 
The impact of flight altitude on performance is quantitatively evaluated in Figure \ref{fig:flyaltitu}, with altitudes categorized into low ($<$30m), mid ($\ge$30m and $\le$70m), and high ($>$70m) tiers. 
For generic MLLMs and our fine-tuned low-altitude models, the performance is generally better at low flight altitudes and the worst at high flight altitudes.
This is primarily attributed to the reduction in object pixel resolution at higher altitudes. As targets occupy fewer pixels, critical morphological and textural features are lost, making semantic recognition and spatial reasoning significantly more difficult. 
Interestingly, the performance of specialized RS-MLLMs often improves as the altitude increases. This suggests that higher altitudes produce visual characteristics that more closely align with the satellite remote sensing imagery distributions on which these models are pre-trained. 
Furthermore, several models achieve their peak performance within the mid-altitude range for specific tasks (\textit{e.g.}, VQA task). This occurs because the mid-altitude perspective provides a sufficiently broad field of view to capture essential environmental context, while simultaneously ensuring that the objects retain a large enough scale. 
}

{
\textbf{Performance biases across target quantities.}
The complexity of low-altitude scenes is further analyzed through performance variance across different target quantities in Figure \ref{fig:objectnumber}, classified as Easy (1-8 objects), Moderate (9-20), and Hard ($>$20). 
In the Target Counting task, all evaluated models experience a sharp performance decline as the object density increases. For example, Qwen2.5-VL drops from $38.15\%$ accuracy in Easy scenarios to a mere $3.17\%$ in Hard scenarios. Even fine-tuned MLLMs still have this counting bottleneck.
However, in the Region Classification task, the impact of target quantities on model performance is very small. This suggests that as long as the accurate position of the object is provided, the MLLM will not be disturbed by other objects within dense environments.
}

{
\textbf{Domain gap between natural/satellite and low-altitude UAV data.}
}
The generic MLLMs have encountered obstacles in effectively processing low-altitude visual data due to the lack of expert knowledge of this domain. However, GeoChat and SkyEyeGPT, which are specifically designed for remote sensing, perform even worse. The results show that the gap between low-altitude images and satellite remote sensing images is larger than that of natural images. This is because low-altitude images are highly consistent with natural images in terms of color representation, spatial resolution, and texture details, with only scene perspectives and field of view being different. In contrast, satellite remote sensing images exhibit significant differences due to the unique imaging mechanism and observation scale.

{
To provide quantitative evidence of the domain gap, we conduct a comparative study on the data of the low-altitude UAV domain, the natural domain, and the satellite RS domain.
We randomly sample 1k images from each of the three domains. Data representing the three domains are randomly sampled, respectively, from our proposed dataset, the ImageNet dataset, and the DIOR dataset. 
As shown in Table \ref{tab:RSuavdis}, we calculate the Fréchet Inception Distance (FID) \cite{NIPS2017_8a1d6947} and Cosine Similarity over 20 random sampling iterations. 
The results quantify that the distance between UAV and satellite RS data is significantly larger than the distance to natural data. The higher cosine similarity scores also quantitatively confirm that the distribution of low-altitude UAV imagery is more similar to that of natural data.
We perform t-SNE clustering to visualize the distribution of these visual features in a 2D space, as shown in Figure \ref{fig:T-SNE}. The UAV images and natural images occupy the same side of the feature space, while satellite images are distributed on the opposite side. There is a significant amount of overlap between the UAV and natural image clusters, suggesting a higher degree of shared semantic and textural characteristics. In contrast, the UAV images and satellite RS images are generally isolated from one another with no significant overlap. 
This visual separation confirms that the domain gap between low-altitude UAV data and satellite RS imagery is far more pronounced than the gap between UAV and natural images.
}

\begin{table*}[t]
\centering
\caption{Ablation studies on the task identifier, data scale, and task level. 
The result of each task corresponds to the average of the out-of-domain tests $\dagger$ in Tables \ref{tab:imageclsVQA} - \ref{tab:regioncap}.
}
\label{tab:ablation_studies}
\begin{tabular}{cccccccccc}
\toprule
 \multirow{3}{*}{\textbf{\makecell[c]{Task \\identifier}}} & \multirow{3}{*}{\textbf{\makecell[c]{Data \\scale}}} &\multicolumn{2}{c}{\textbf{Task level}}  & \multicolumn{3}{c}{\textbf{Image-level} $\dagger$} & \multicolumn{3}{c}{\textbf{Region-level} $\dagger$}  \\ \cmidrule(lr){3-4} \cmidrule(lr){5-7} \cmidrule(lr){8-10}
 & &  Image  & Region   & Cls  & VQA & Count & R.VQA & R.Cls & VG (\%)\\
 & & level  & level  & (Acc(\%)) & (Acc(\%)) & (Acc(\%)) & (Acc(\%)) & (Acc(\%))  &(Acc@0.5) \\ 
 \midrule
 \multicolumn{10}{c}{{\textbf{LLaVA1.5-UAV baseline}}} \\
\midrule
\XSolidBrush & 100\%  & \Checkmark & \Checkmark & 81.82 &94.20 & 23.51 & 46.78&93.71 &3.04\\
 \Checkmark & 100\%  &\Checkmark & \Checkmark & {\textbf{83.00}} &{\textbf{98.94}}&\underline{24.59} & {\textbf{48.48}}& {\textbf{96.32}}  &{\textbf{3.73}}\\
\Checkmark & 50\%  & \Checkmark & \Checkmark & 81.47 &93.48 & 23.33 &45.37 &90.97&2.22\\
\Checkmark & 10\%  & \Checkmark & \Checkmark & 78.28&87.74 & 18.58 & 40.54 & 79.51 &1.13\\
\Checkmark &-  &  \Checkmark & \XSolidBrush & \underline{82.23} &\underline{97.11}  &{\textbf{24.69}} & 36.91 & 56.48 &0.14 \\
\Checkmark &-  &  \XSolidBrush & \Checkmark & 76.91 &90.68 &17.14 & \underline{48.17} & \underline{95.11}&\underline{3.71} \\
\midrule
 \multicolumn{10}{c}{{\textbf{MiniGPTv2-UAV baseline}}} \\
\midrule
\XSolidBrush & 100\%  & \Checkmark & \Checkmark &84.79 &95.45 & 23.18 & 48.77&88.27 &\underline{27.81}\\
\Checkmark & 100\%  & \Checkmark & \Checkmark & \underline{85.69} &{\textbf{97.35}}&{\textbf{25.46}} & {\textbf{50.65}}& \textbf{90.62}  &{\textbf{28.27}}\\
 \Checkmark & 50\%  &\Checkmark & \Checkmark & 82.41 &95.69 &22.23 &48.18&87.12&24.67\\
\Checkmark &10\%  &  \Checkmark & \Checkmark & 79.32&88.60 & 18.51 & 41.88 &80.53 &17.65\\
\Checkmark & -  & \Checkmark & \XSolidBrush & {\textbf{86.97}} &\underline{96.20}  & \underline{{24.97}} & 35.74 & 55.13&5.23 \\
\Checkmark & -  & \XSolidBrush & \Checkmark & 77.18 &85.24 &16.85 & \underline{49.82} & \underline{89.70}&{27.73} \\
\bottomrule
\end{tabular}
\end{table*}

 \begin{figure*}[t]
  \centering
  \includegraphics[width=0.75\linewidth]{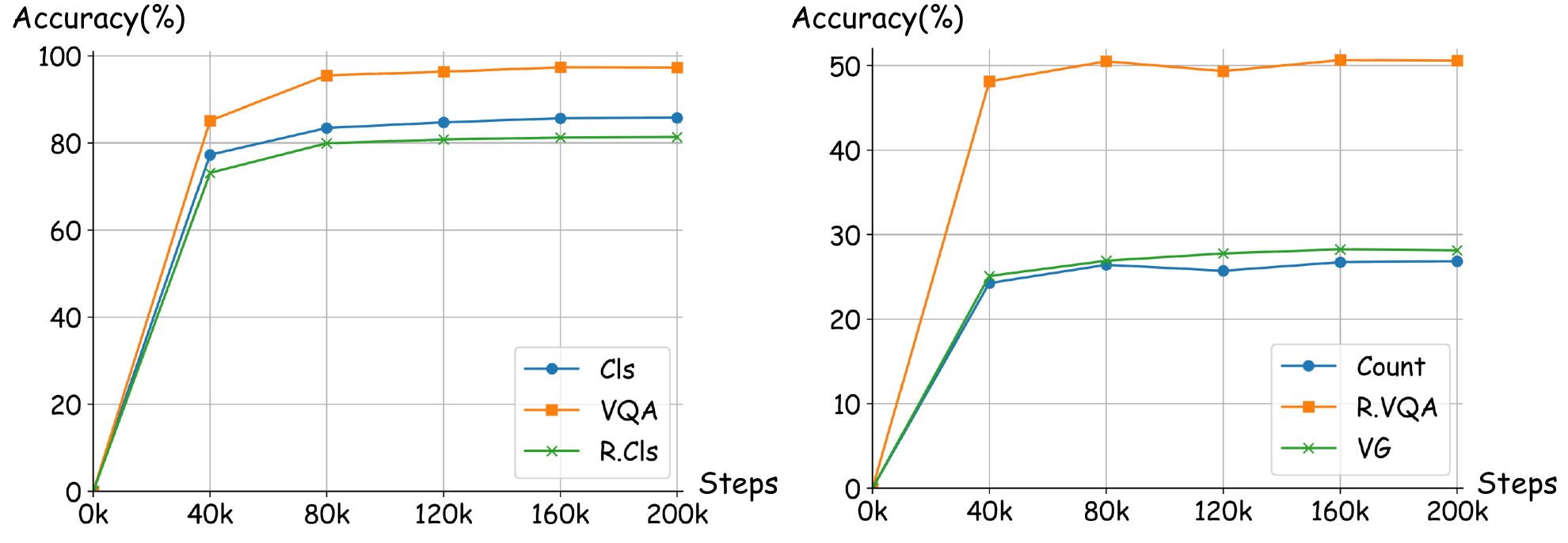}
  \caption{
  {The zero-shot performance of the first 200k training steps during fine-tuning of MiniGPT-v2 on UAVIT-1M.}
  }
  \label{fig:MiniGPTUAV}
\end{figure*}

\subsection{Ablation Studies}
In this section, we analyze the impact of the task identifier, data scale, and task level on performance by ablation studies. We test the baseline LLaVA1.5-UAV and baseline MiniGPTv2-UAV, as shown in Table \ref{tab:ablation_studies}.
We observe that: (1) Adding task identifiers reduces the ambiguity among various tasks, which benefits the overall performance and multi-task learning efficiency.
(2) The performance continues to decline as the instruction-tuning data scale decreases.
(3) Multi-granularity tasks can effectively promote each other and enhance the performance of simple and complex tasks.

{
To investigate the impact of instruction-tuning scale on model performance, we fine-tune the MiniGPT-v2 on the UAVIT-1M dataset with different training steps. 
We monitor the zero-shot accuracy of MiniGPT-v2 at intervals of 40k steps during the first 200k steps of fine-tuning on UAVIT-1M.
As illustrated in Figure \ref{fig:MiniGPTUAV}, the learning dynamics exhibit distinct characteristics across different tasks. All tasks demonstrate a steep performance improvement in the initial stage.
Performance trends diverge as training progresses.
Semantic understanding tasks (\textit{e.g.}, Cls, VQA, and R.Cls) rapidly reach high performance ($>80\%$) after approximately 80k steps. 
In contrast, complex reasoning and localization tasks (\textit{e.g.}, Count, R.VQA, VG) show slower convergence and remain stable at lower accuracy thresholds.
This analysis suggests that while basic semantic alignment is achieved rapidly with our UAVIT-1M dataset, fine-grained counting and localization tasks require specialized model design or training paradigms to break through performance bottlenecks.
}

\begin{figure*}[!ht]
  \centering
  \includegraphics[width=0.9\linewidth]{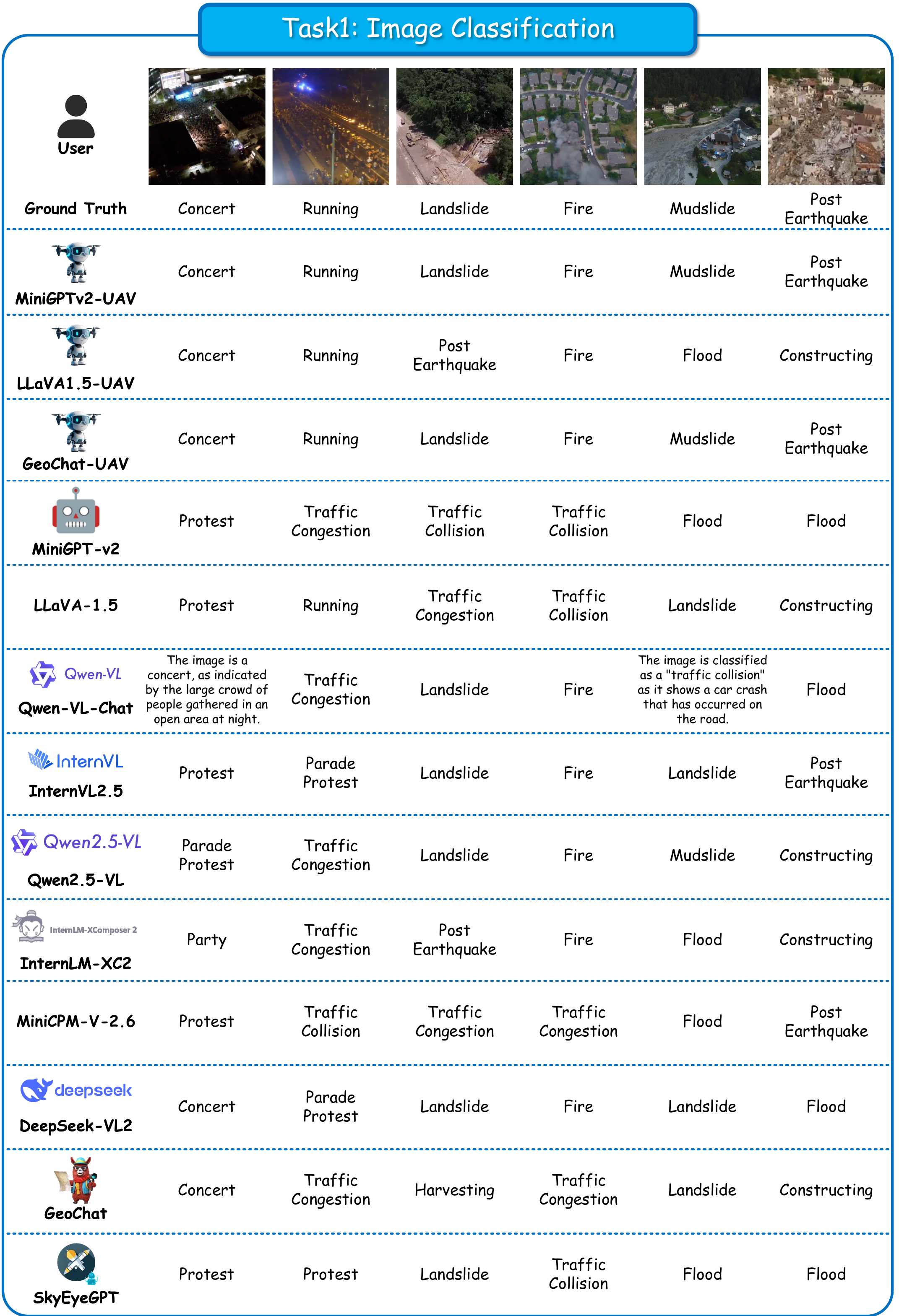}
  \caption{
  Qualitative comparisons in the task of image classification.
  }
  \label{vis_cls}
\end{figure*}

\begin{figure*}[!ht]
  \centering
  \includegraphics[width=0.9\linewidth]{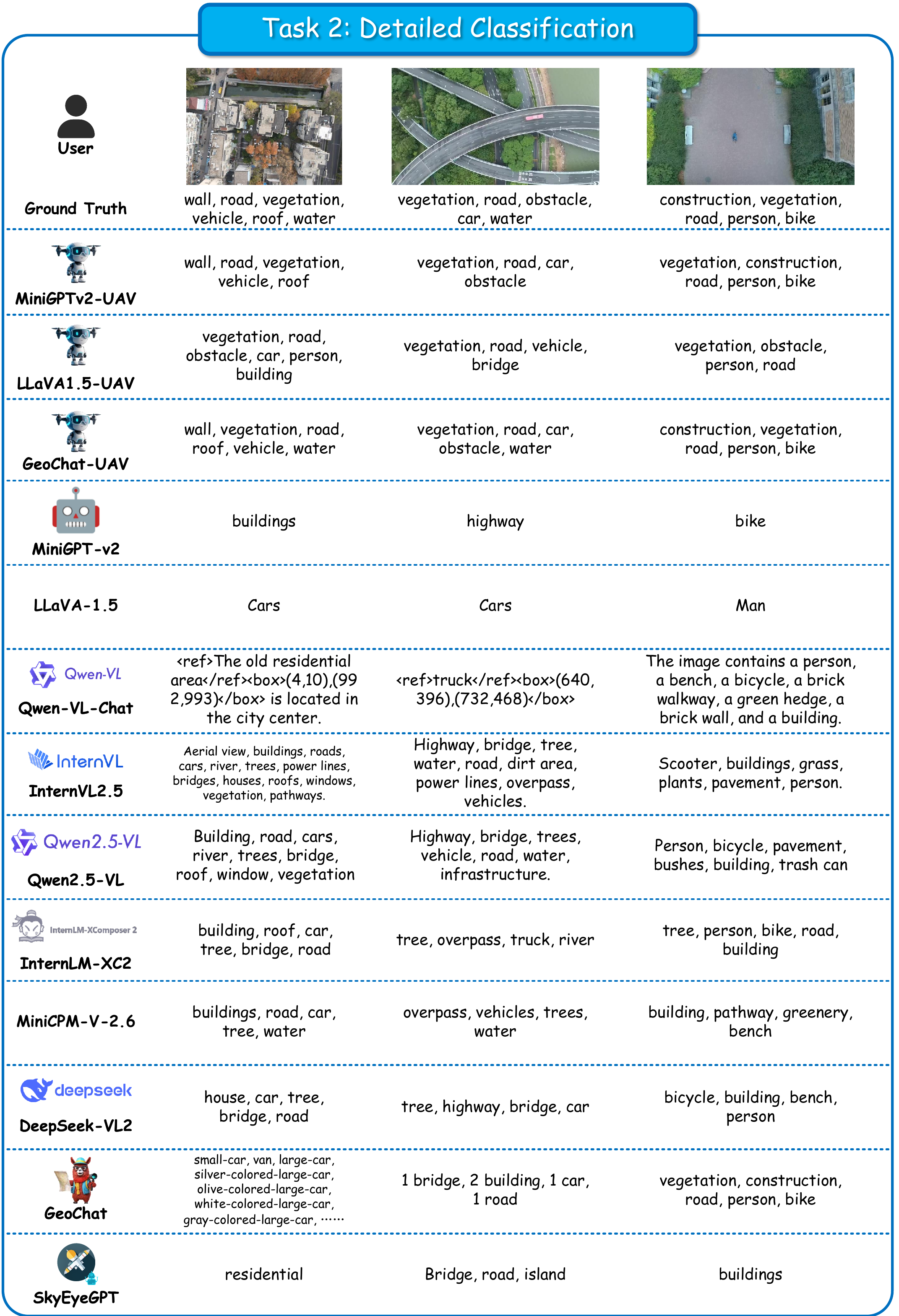}
  \caption{
  Qualitative comparisons in the task of detailed classification.
  }
  \label{vis_dcls}
\end{figure*}

\begin{figure*}[!ht]
  \centering
  \includegraphics[width=0.88\linewidth]{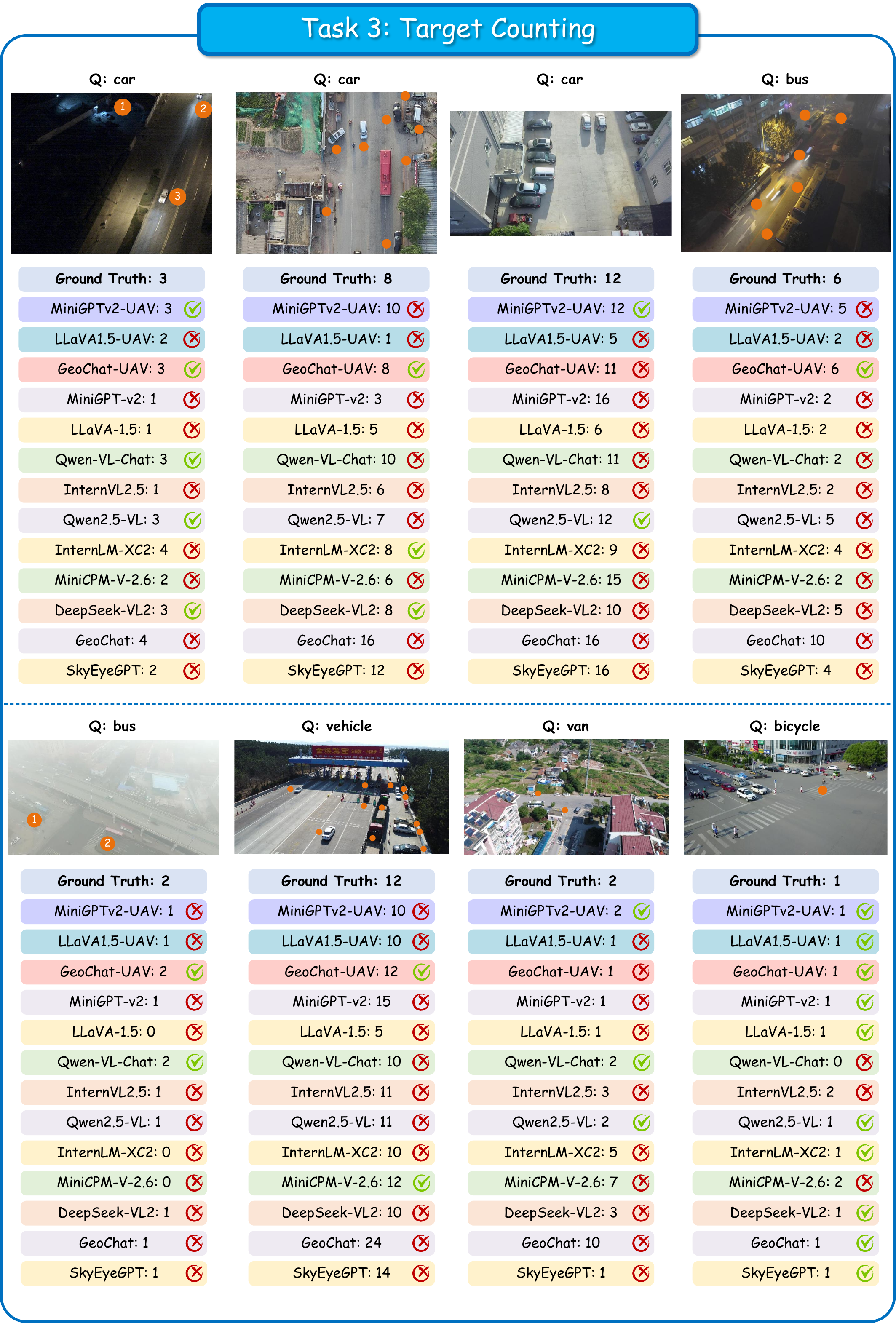}
  \caption{
  Qualitative comparisons in the task of target counting.
  }
  \label{vis_count}
\end{figure*}

\begin{figure*}[!ht]
  \centering
  \includegraphics[width=0.88\linewidth]{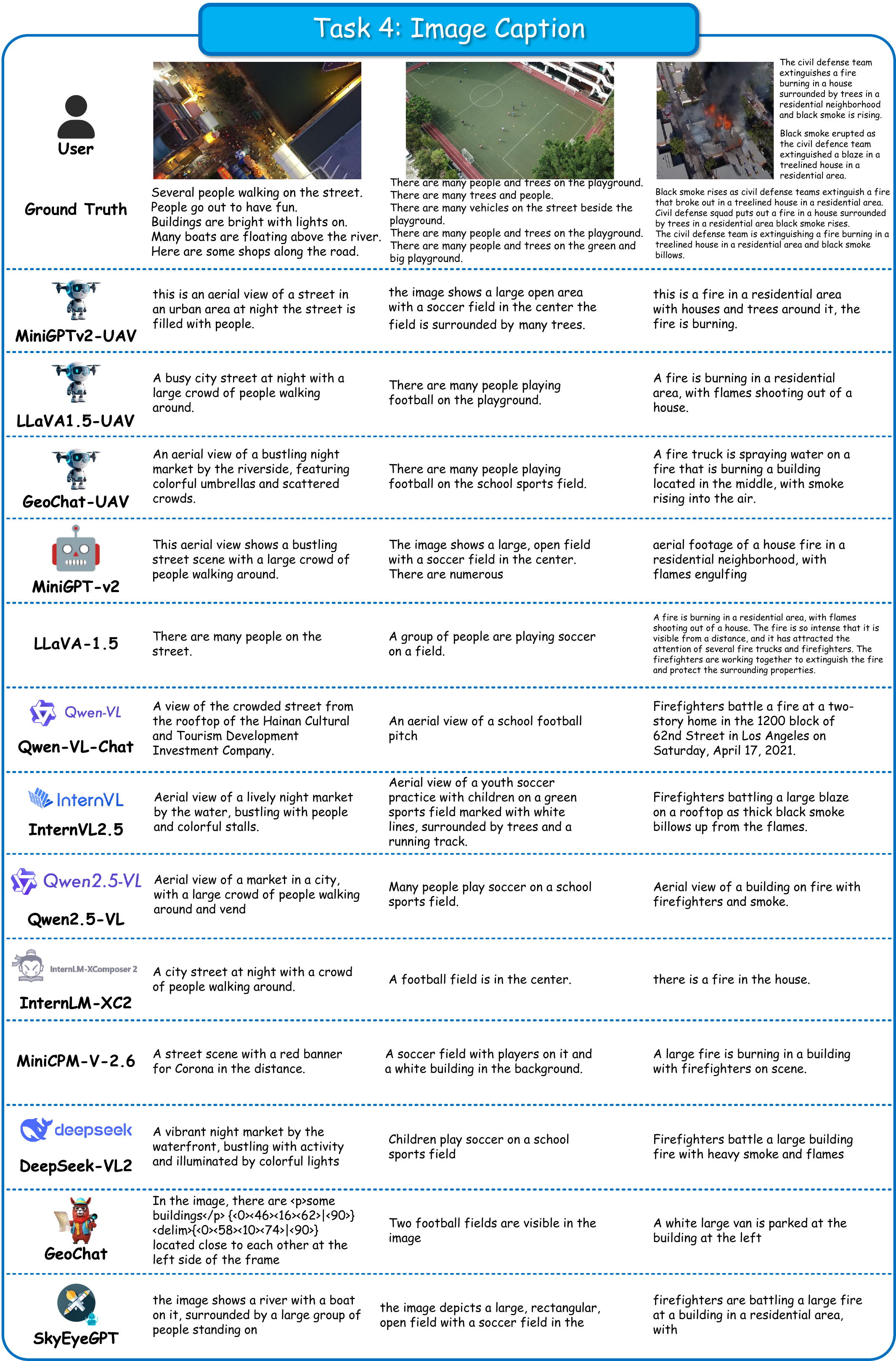}
  \caption{
  Qualitative comparisons in the task of image captioning.
  }
  \label{vis_cap}
\end{figure*}

\begin{figure*}[!ht]
  \centering
  \includegraphics[width=0.9\linewidth]{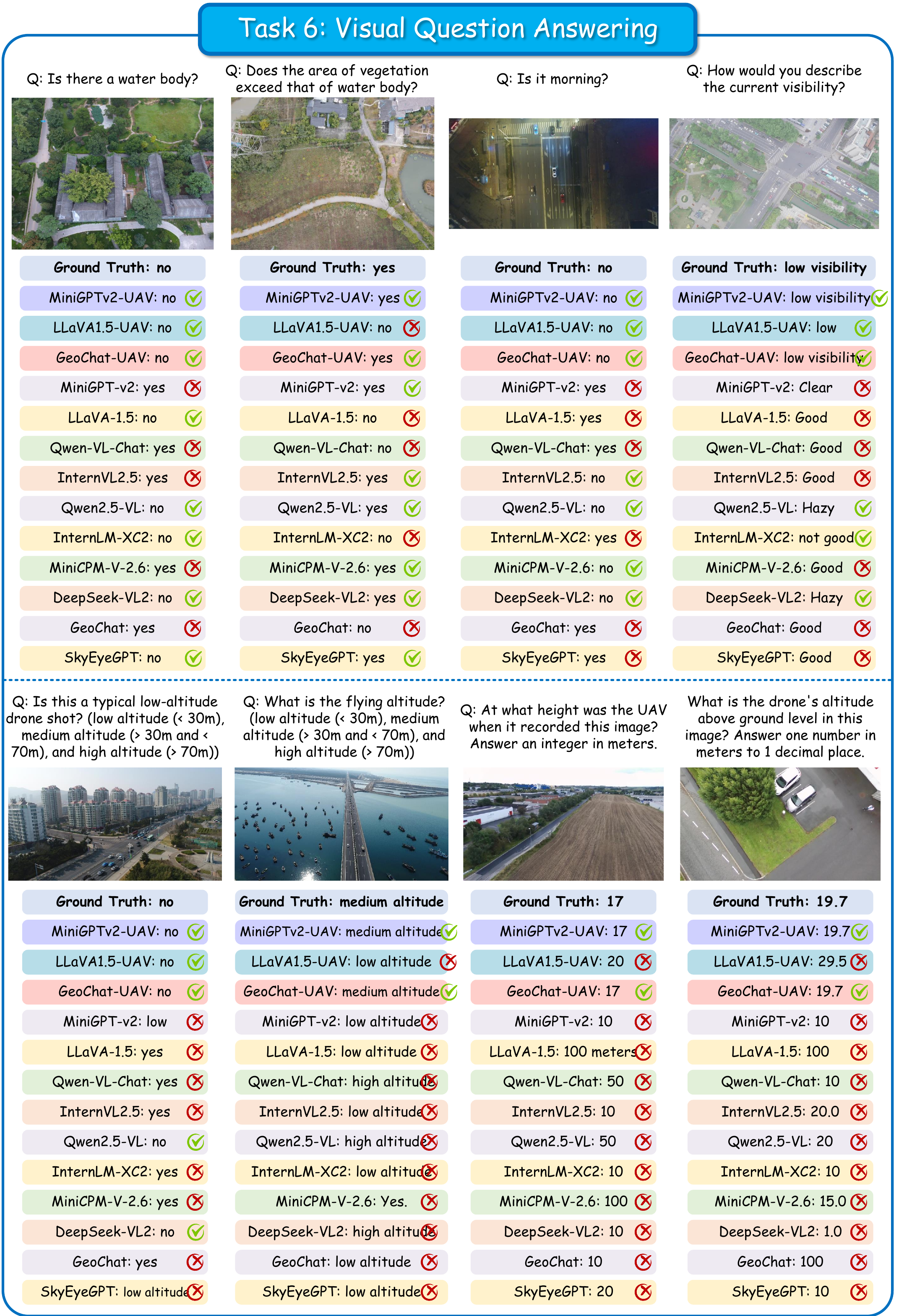}
  \caption{
  Qualitative comparisons in the task of visual question answering.
  }
  \label{vis_vqa}
\end{figure*}

\begin{figure*}[!ht]
  \centering
  \includegraphics[width=0.9\linewidth]{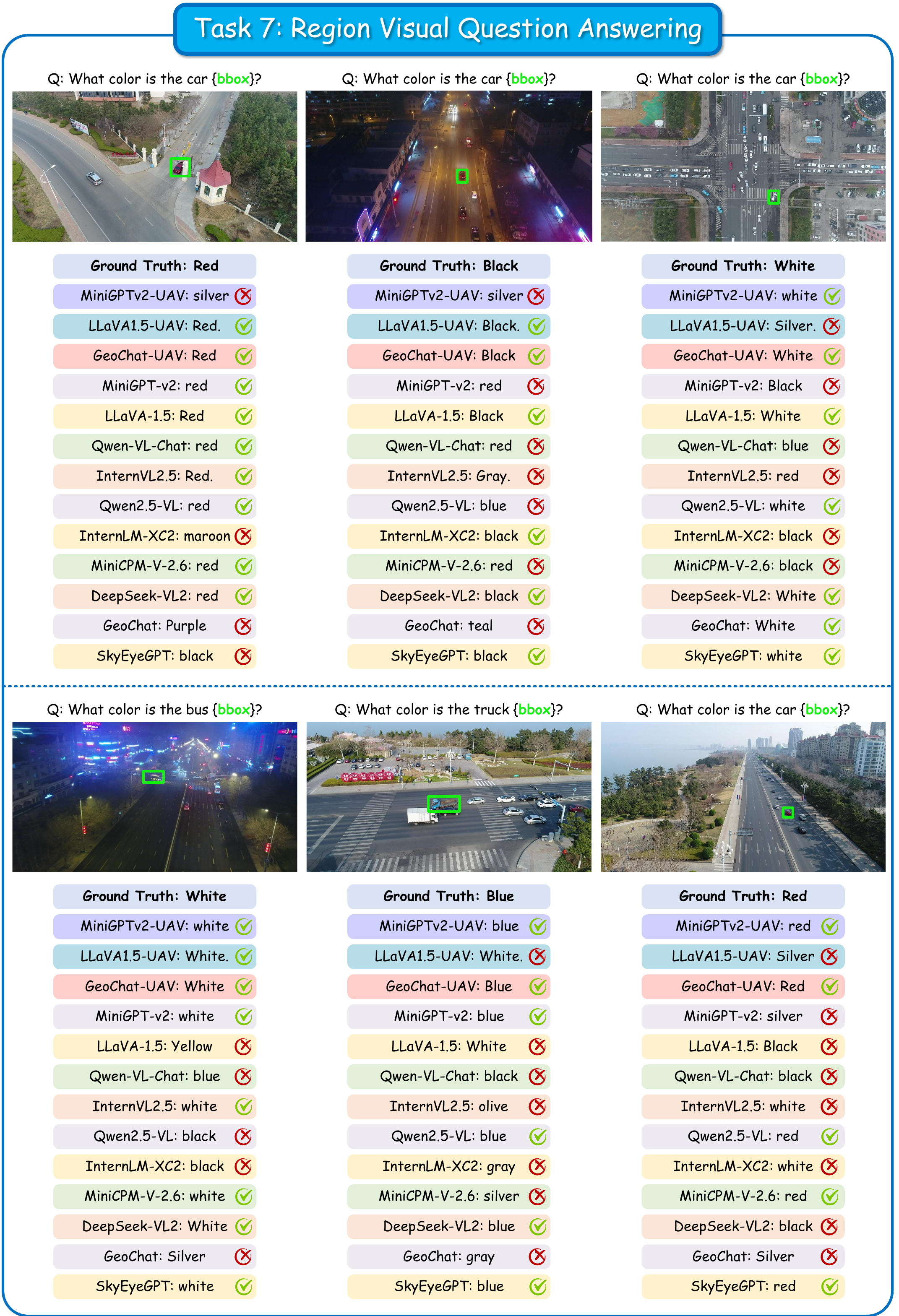}
  \caption{
  Qualitative comparisons in the task of region visual question answering.
  }
  \label{vis_rvqa}
\end{figure*}

\begin{figure*}[!ht]
  \centering
  \includegraphics[width=0.88\linewidth]{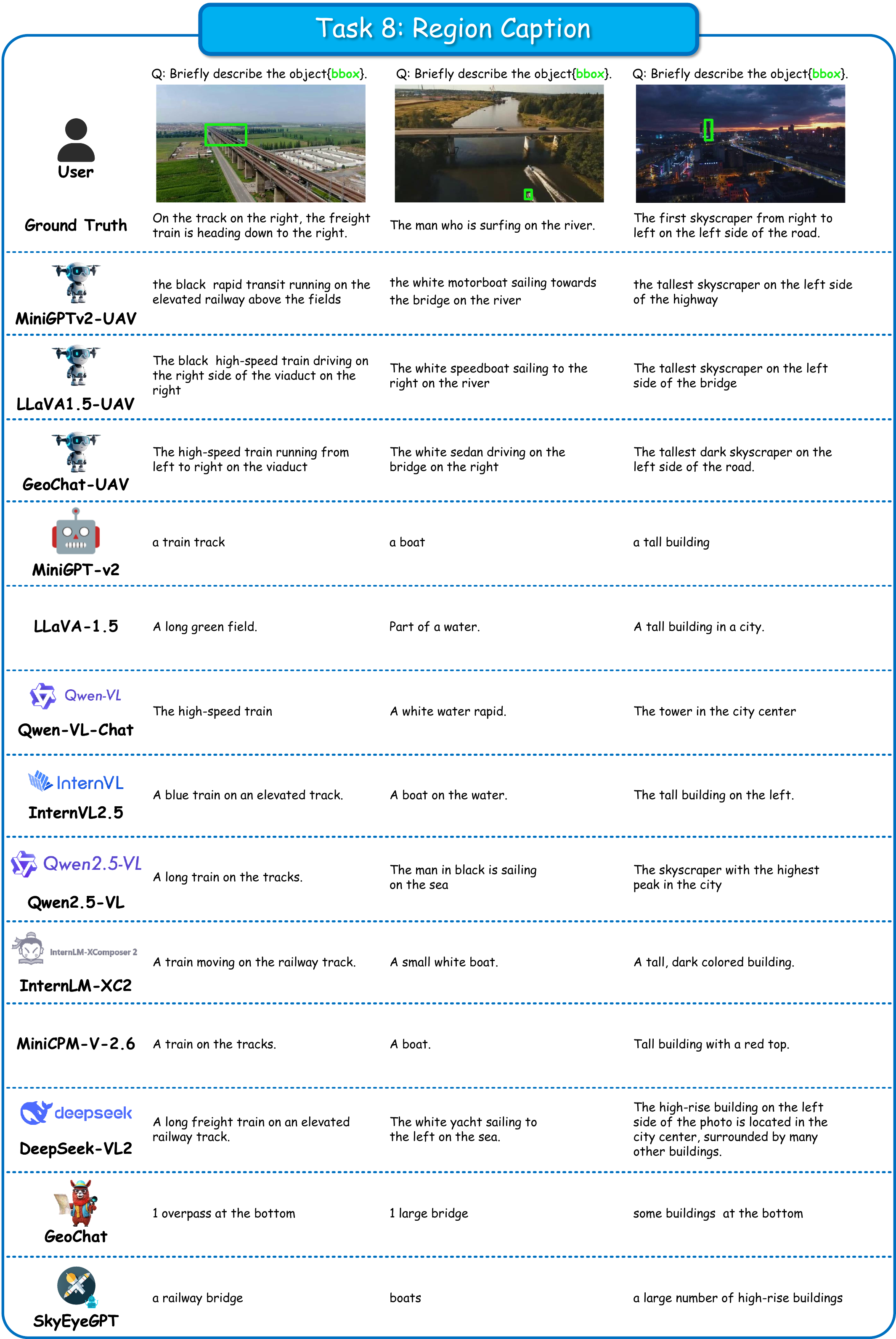}
  \caption{
  Qualitative comparisons in the task of region caption.
  }
  \label{vis_rcap}
\end{figure*}

\begin{figure*}[!ht]
  \centering
  \includegraphics[width=0.88\linewidth]{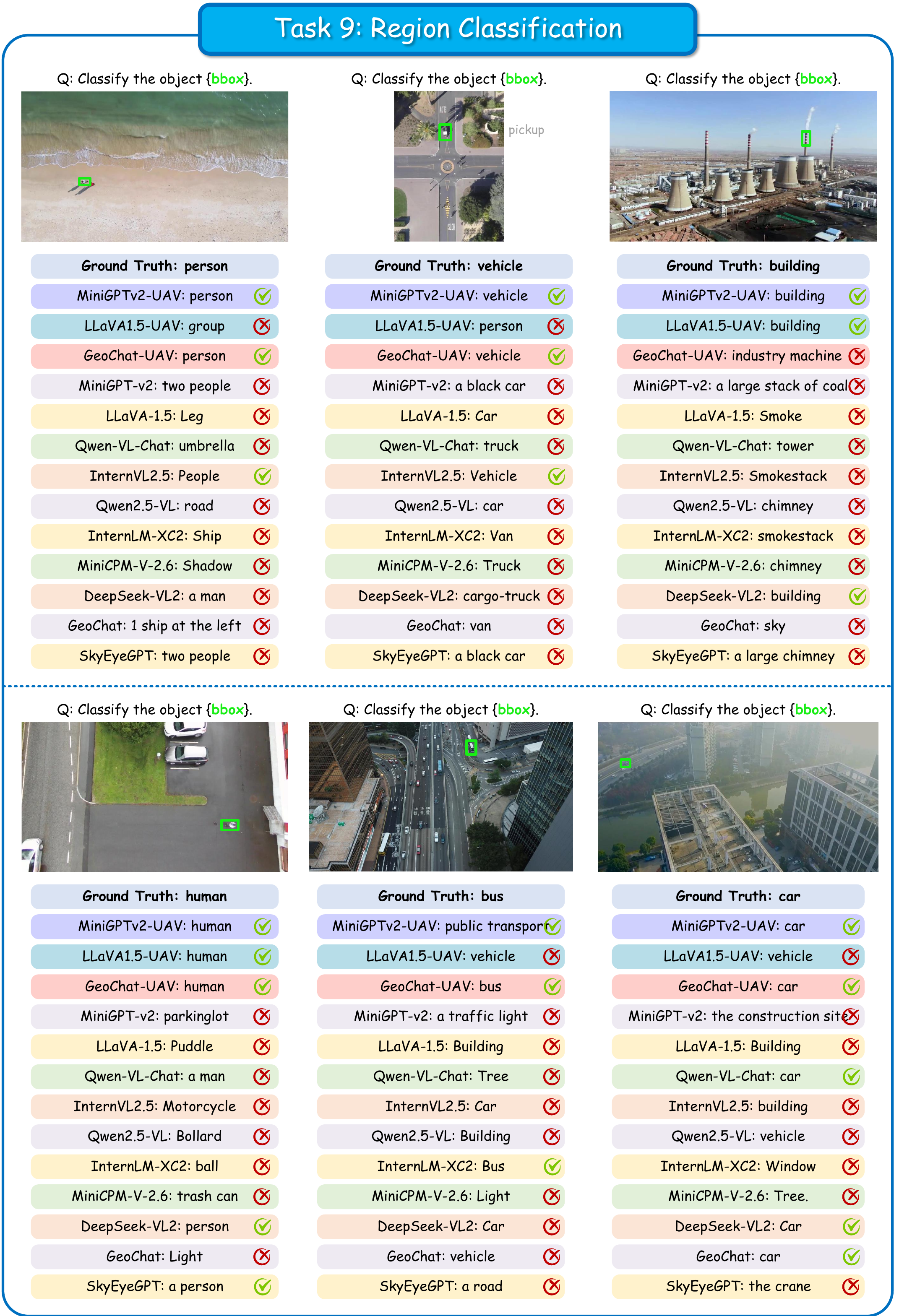}
  \caption{
  Qualitative comparisons in the task of region classification.
  }
  \label{vis_rcls}
\end{figure*}

\begin{figure*}[!ht]
  \centering
  \includegraphics[width=0.95\linewidth]{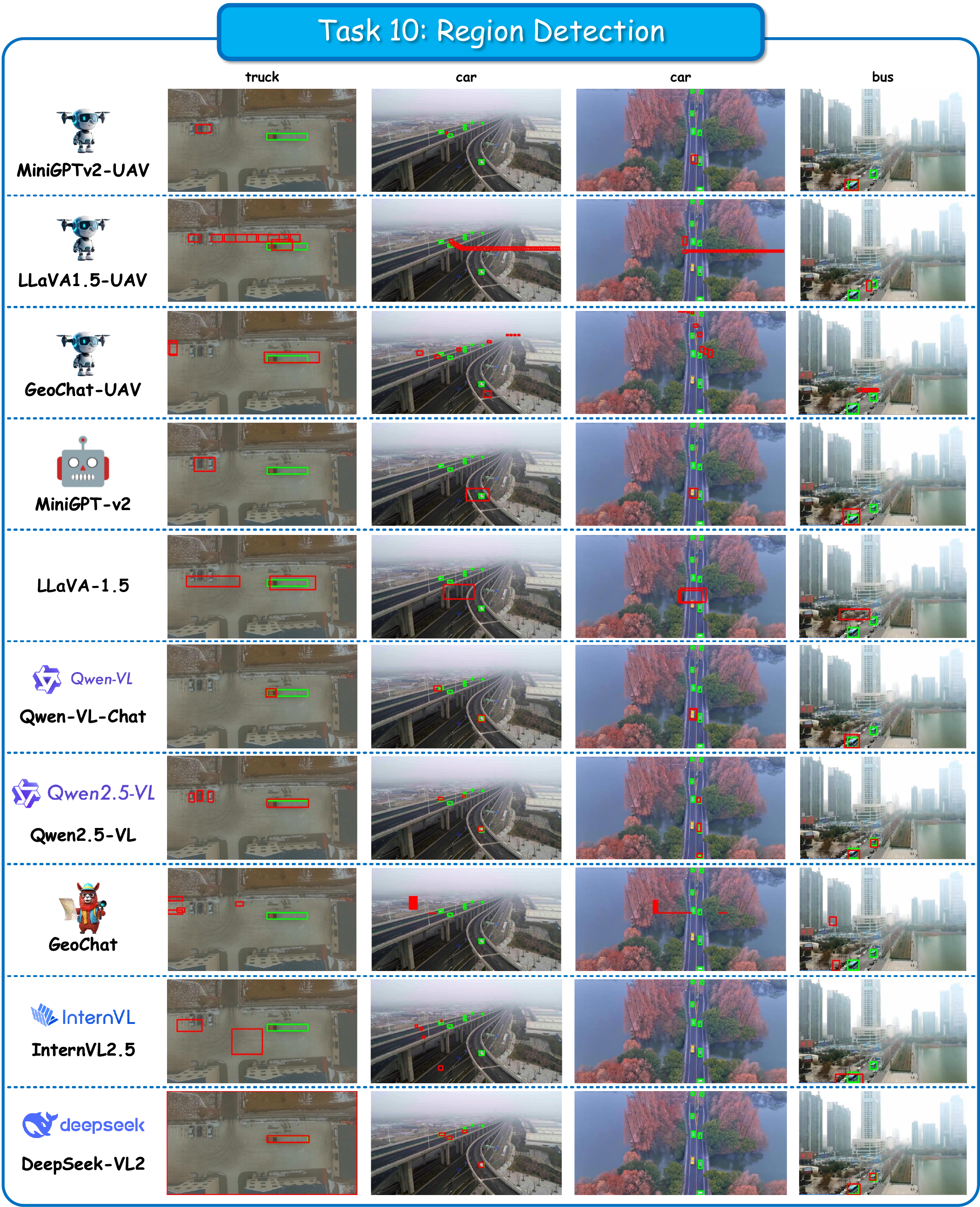}
  \caption{
  Qualitative comparisons in the task of region detection.
  }
  \label{vis_det}
\end{figure*}

\begin{figure*}[!ht]
  \centering
  \includegraphics[width=0.95\linewidth]{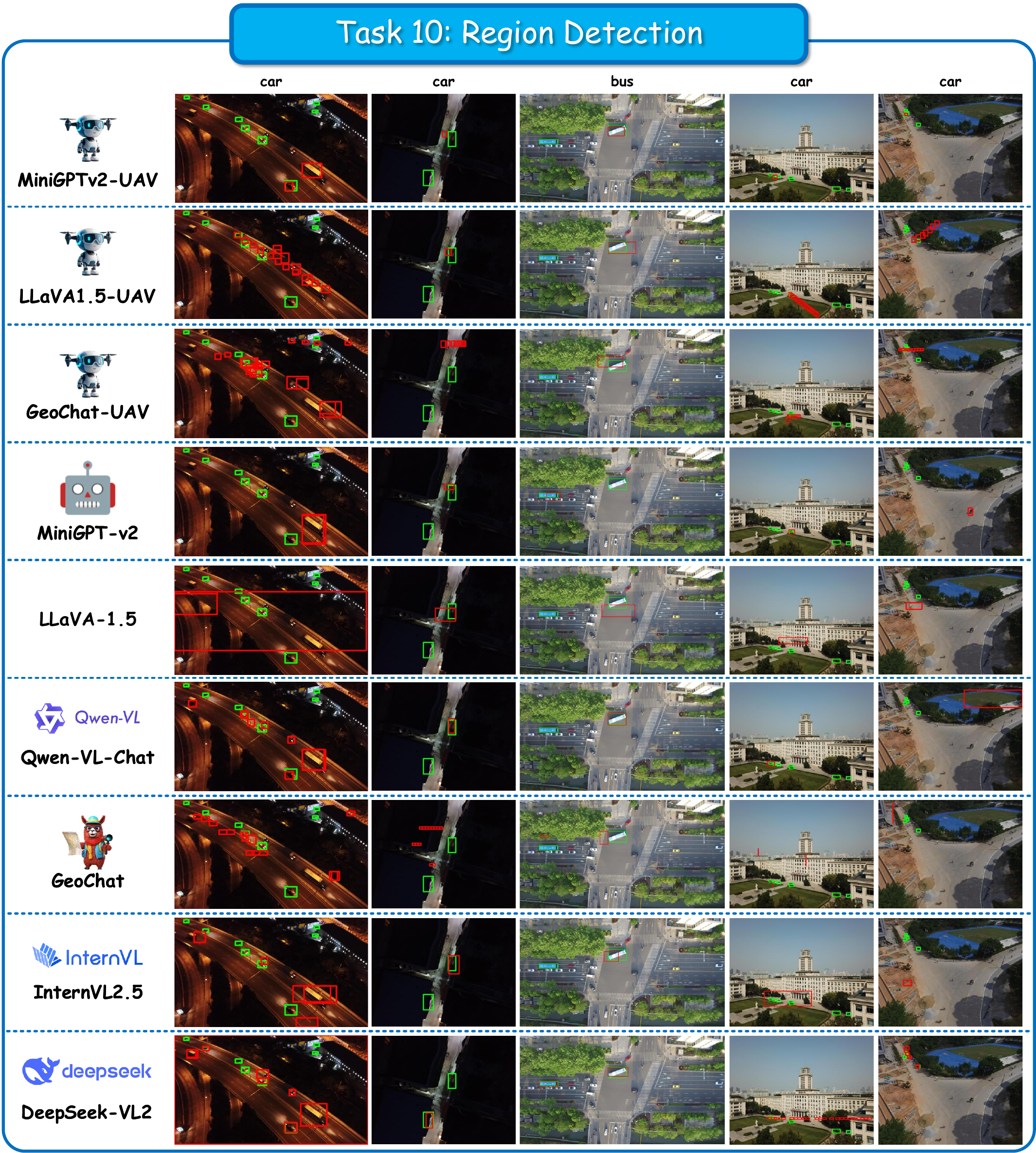}
  \caption{
  Qualitative comparisons in the task of region detection.
  }
  \label{vis_det2}
\end{figure*}

\begin{figure*}[!ht]
  \centering
  \includegraphics[width=0.88\linewidth]{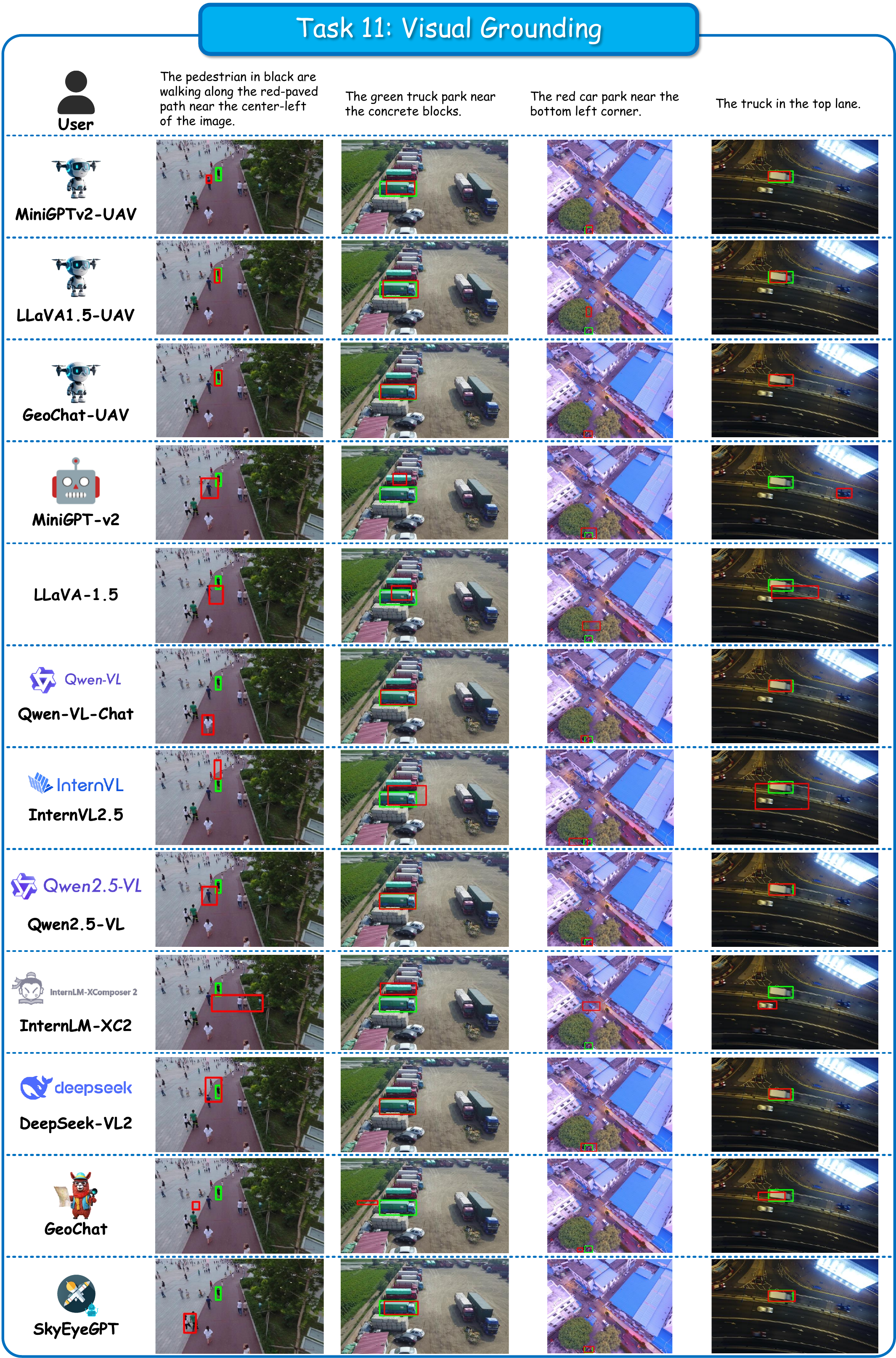}
  \caption{
  Qualitative comparisons in the task of visual grounding.
  }
  \label{vis_rvg}
\end{figure*}

\subsection{Vision-Language Fusion and Alignment Challenges}
{
Although quantitative results demonstrate the performance gap of MLLMs in UAV scenarios, it is crucial to analyze the underlying mechanisms affecting visual-language fusion. Unlike natural images or satellite RS images, low-altitude data introduces unique fusion and alignment challenges:
}

{
\textbf{Spatial-Semantic Alignment}: The core challenge lies in the perspective gap. Most general MLLMs utilize visual encoders (\textit{e.g.}, CLIP) pre-trained on internet-scale natural images dominated by frontal views. However, low-altitude UAV images feature diverse pitch angles (oblique views). This geometric deformation causes a distribution shift in visual features. Consequently, the multimodal projection layer (MLP/Linear) struggles to map these "unfamiliar" visual tokens to the semantic space of the LLM, leading to alignment failures where the model sees the pixels but cannot associate them with the correct object concept (\textit{e.g.}, failing to recognize a "person" from a top-down view).
}

{
\textbf{Weather-Induced Modality Imbalance}: Our evaluation on datasets like HazyDet reveals that adverse weather reduces performance. Fog or low light degrades the contrast and saliency of visual features.
During the fusion process, if attention mechanisms in MLLMs assign higher weights to these visual tokens, the unclear object information often fails to achieve true multimodal understanding.
If MLLMs assign higher weights to textual tokens, they tend to over-rely on language priors, thereby resulting in hallucinations.
}

{
\textbf{Altitude-Related Scale Variations}: The scale of the target varies with the flight altitude. As demonstrated in Figure 9, performance degrades significantly at higher altitudes. In V-L fusion, visual encoders typically process images in fixed-size patches (\textit{e.g.}, $14 \times 14$ or $16 \times 16$). In high-altitude UAV images, objects of interest often occupy fewer pixels than a single patch. This results in critical feature loss during the encoding stage. When these empty or blurred visual tokens are fused with text instructions, the LLM lacks sufficient visual evidence to generate precise responses.
}

\subsection{Qualitative Comparisons}
Please see Figure \ref{vis_cls} - Figure \ref{vis_rvg}.
Due to space limitations, please refer to the project homepage for the complete results across all tasks.

\section{Future Directions}
In this section, we discuss potential directions for future research.

{
\textbf{Novel Model Architecture.}
To address the inherent challenges of low-altitude vision, such as scale variations and small objects, novel architecture designs should be carried out in the vision encoder and multimodal alignment layer.
A promising direction involves constructing a robust and hybrid visual perception system by integrating various pre-trained models, such as CLIP ViT, EVA ViT, DINOv2 ViT, and CLIP ConvNeXt-L. By leveraging their complementary strengths, future architectures should focus on mining multi-scale and multi-layer visual features. This would enable the extraction of rich semantic information while preserving the fine-grained spatial details.
Enhancing the modality alignment layer with adaptive fusion or cross-scale attention mechanisms will further ensure that these multi-level visual cues are precisely mapped into the linguistic subspace.
Specifically, the Fourier Transform or Frequency Decoupling can be utilized to separate high-frequency structural details from low-frequency global information. 
The integration of Wavelet Transform or Information Entropy could further enable the MLLM to selectively attend to information-rich regions. 
Additionally, applying Manifold Learning to the alignment may facilitate a more topologically consistent mapping between visual and linguistic embeddings. 
By fully mining multi-scale and multi-layer features, future UAV-specialized MLLMs can achieve more robust semantic comprehension and precise spatial reasoning in complex, dynamic low-altitude missions.
}

{
\textbf{Reinforcement Fine-Tuning Strategy.}
Instruction fine-tuning belongs to the supervised learning paradigm and relies on a large amount of labeled data.
Inspired by the recent success of the Group Relative Policy Optimization (GRPO) paradigm introduced by DeepSeek, which significantly enhances the reasoning capabilities of LLMs, realizing chain-of-thought (CoT) reasoning in complex low-altitude UAV scenarios will have great potential.
Future research could focus on integrating CoT with reinforcement learning to stimulate slow thinking in low-altitude MLLMs and agents. 
By defining reward functions based on the precision of coordinate grounding and the accuracy of classification, counting, and VQA, models can be incentivized to self-correct and explore optimal reasoning paths. 
Such a reinforcement-based fine-tuning strategy would enable UAV-specialized MLLMs to effectively handle the complex reasoning and visual ambiguity inherent in low-altitude tasks.
}

\textbf{Region-level Perception.}
Our UAVBench benchmark poses significant challenges to current MLLMs. The performance of the model fine-tuned on UAVIT-1M has been significantly improved. 
However, for difficult region-level tasks, the performance is limited. Despite their potential, current MLLMs face challenges in fine-grained region-level tasks, particularly in grounding or detecting objects.
In the future, we plan to propose specific designs to enhance low-altitude MLLMs' region perception ability.

\textbf{Performance Stability.}
Advanced MLLMs and low-altitude MLLMs exhibit certain instability in performance on different tasks. The capabilities of MLLMs are relatively mixed, and their advantages and disadvantages vary.
Therefore, in the future, low-altitude MLLMs should improve the performance balance of different granularities and tasks, and become all-rounders that achieve the best performance in every task.

\textbf{Complementarity of Generalists and Specialists.} 
Future work can combine the universality of MLLMs with the precision of specialist models. MLLMs offer the significant advantage of performing a wide array of tasks through simple instructions. However, their performance currently lags far behind that of specialists for specific tasks. For example, detectors like YOLO or DETR can achieve 40\% to 50\% of AP@0.5 on the VisDrone-2019-DET dataset. This highlights a crucial trade-off between versatility and precision.

\textbf{Support Modality.}
Currently, UAVBench and UAVIT-1M primarily focus on RGB images. In practical applications, UAVs are also often equipped with other types of visual sensors. In future work, we aim to enhance UAVBench and UAVIT-1M by merging annotations from various other multimodal visual data, including infrared images, SAR images, multi- and hyperspectral images, and temporal datasets.

\section{Conclusion}
\label{conclu}
In this work, we introduce \textbf{UAVBench}, a comprehensive and challenging vision-language benchmark for low-altitude UAV image-level and region-level understanding.
Our comprehensive evaluation of 11 advanced MLLMs reveals limited performance.
To address this limitation and facilitate the training of low-altitude MLLMs, we further propose \textbf{UAVIT-1M}, the instruction-tuning dataset comprising 1.24 million diverse instructions, 789k multi-scene low-altitude UAV images, about 2,000 types of spatial resolutions, and 11 distinct image and region-level tasks.
We also demonstrate that fine-tuning MLLMs on our UAVIT-1M improves the performance across various tasks on the UAVBench benchmark. 
Our UAVBench and UAVIT-1M will be beneficial to the research community and foster future advancements in low-altitude intelligence and human-drone interaction.

In light of the fact that the era of low-altitude economy and the field of spatial-temporal intelligence has just begun, we provide comprehensive datasets for evaluating and developing low-altitude MLLMs. 
This is the first to support the training and evaluation of advanced MLLMs, and boost their ability to tackle complex low altitude real-world scenarios. 
Low-altitude MLLMs can enhance the intelligence level of UAVs applied in various industries, such as emergency rescue, industrial inspection, urban governance, traffic management, agricultural environmental protection, and other fields.

%

	
\ifCLASSOPTIONcaptionsoff
\newpage
\fi

\footnotesize
\bibliographystyle{IEEEtranN}
\bibliography{main}

\end{document}